\pdfoutput=1

\documentclass[11pt]{article}

\usepackage[preprint]{acl}

\usepackage{times}
\usepackage{latexsym}

\usepackage[T1]{fontenc}

\usepackage[utf8]{inputenc}

\usepackage{microtype}

\usepackage{inconsolata}

\usepackage{hyperref}
\usepackage{url}
\usepackage{enumitem}
\usepackage{graphicx}
\usepackage[newfloat,frozencache,cachedir=.]{minted}
\usepackage{amsmath}
\usepackage{color}
\usepackage{xcolor}
\usepackage{comment}
\usepackage{tikz}
\usetikzlibrary{shapes,arrows,positioning}
\usepackage{booktabs}
\usepackage{listings}
\usepackage{caption}
\usepackage{subcaption}
\usepackage{algorithm}
\usepackage{algpseudocode}
\usepackage{alltt}
\usepackage{adjustbox}
\usepackage{array}

\newcommand{\method}{Re-Tuning}

\title{\method: Overcoming the Compositionality Limits of Large Language Models with Recursive Tuning}

\author{Eric Pasewark\textsuperscript{1*}, Kyle Montgomery\textsuperscript{1*}, Kefei Duan\textsuperscript{1}, Dawn Song\textsuperscript{2}, Chenguang Wang\textsuperscript{1}$^\dagger$\\
\textsuperscript{1}Washington University in St. Louis, \textsuperscript{2}UC Berkeley\\
\texttt{\{eric.pasewark, kylemontgomery, d.kefei, chenguangwang\}@wustl.edu}\\ \texttt{dawnsong@berkeley.edu}}

\begin{document}
\maketitle
\def\thefootnote{$^*$}\footnotetext{Equal contribution.}
\def\thefootnote{$^\dagger$}\footnotetext{Corresponding author.}
\def\thefootnote{$\color{white} ^*$}\footnotetext{The code is available at \url{https://github.com/Pasewark/ReTuning}.}
\begin{abstract}
We present a new method for large language models to solve compositional tasks. Although they have shown strong performance on traditional language understanding tasks, large language models struggle to solve compositional tasks, where the solution depends on solving smaller instances of the same problem. We propose a natural approach to solve compositional tasks recursively. Our method, \method , tunes models to break down a problem into subproblems, solve those subproblems, and combine the results. We show that our method significantly improves model performance on three representative compositional tasks: integer addition, dynamic programming, and parity. Compared to state-of-the-art methods that keep intermediate steps towards solving the problems, \method\ achieves significantly higher accuracy and is more GPU memory efficient.
\end{abstract}
\section{Introduction}
\begin{figure*}[t]
    \centering
    \begin{subfigure}[b]{\textwidth}
        \centering
        \includegraphics[width=\textwidth]{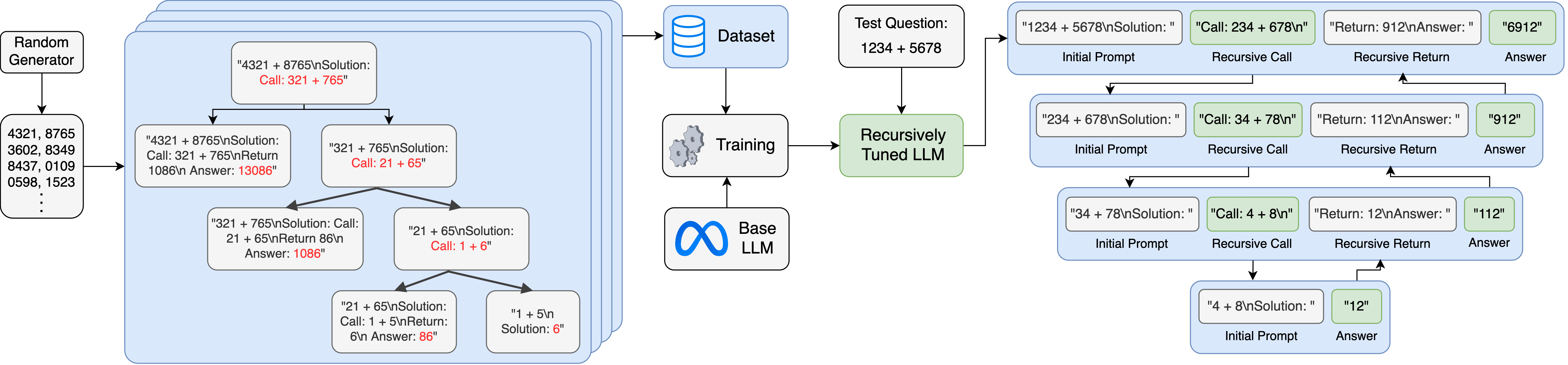}
        \caption{\method\ pipeline.}
        \label{fig:main}
    \end{subfigure}\\
    \par\bigskip
    \begin{subfigure}[b]{0.54\textwidth}
        \centering
        \includegraphics[width=\textwidth]{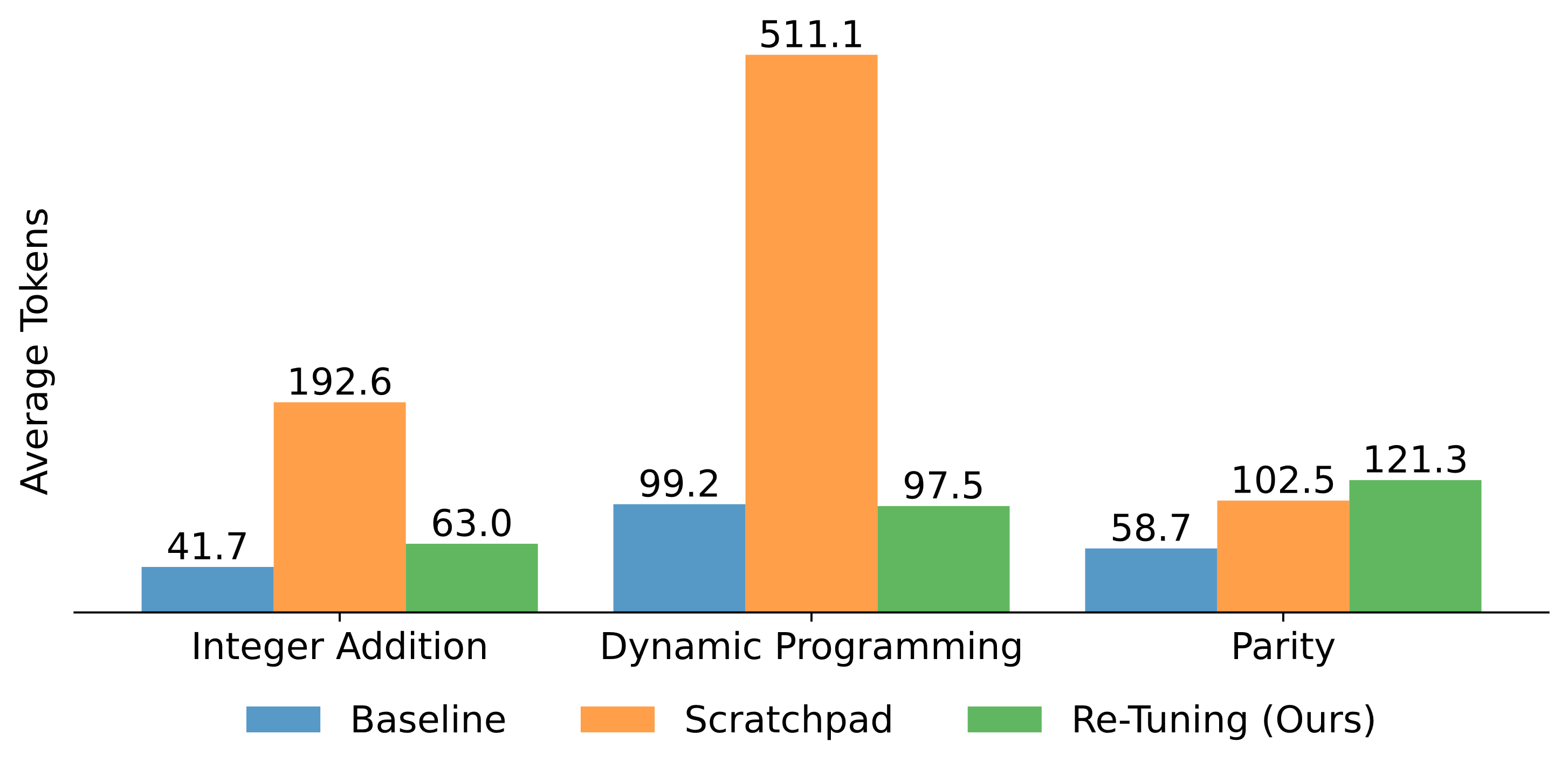}
        \caption{Average tokens per context.}
        \label{fig:context}
    \end{subfigure}
    \hfill
    \begin{subfigure}[b]{0.44\textwidth}
        \centering
        \includegraphics[width=\textwidth]{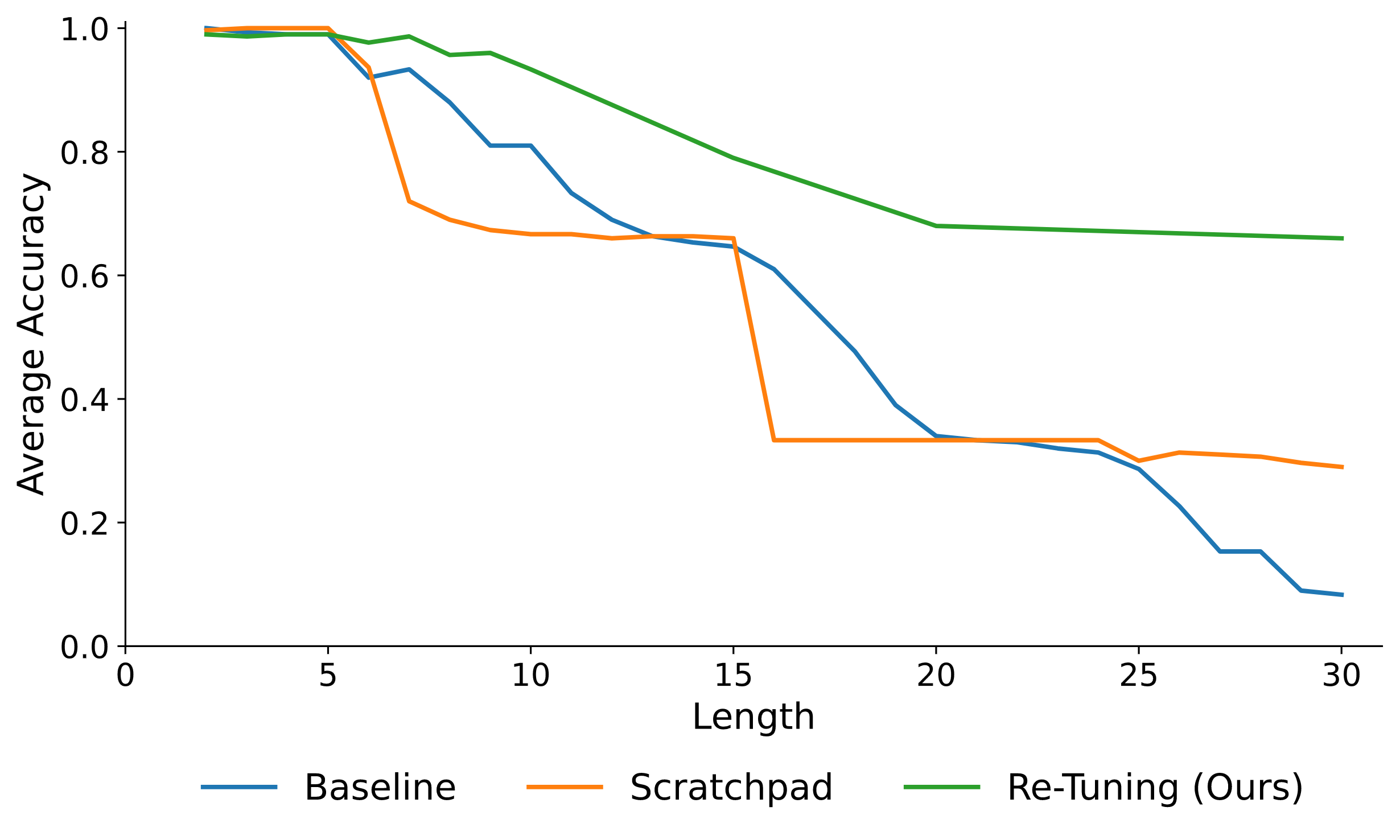}
        \caption{Average performance vs. problem length.}
        \label{fig:agg}
    \end{subfigure}
    \caption{Summary of our approach and results. Top: Our \method\ pipeline generates and processes all the recursive subproblems for each randomly generated problem instance in order to train the base LLM. For a new question, our \method\ pipeline allows the model to call itself on a subproblem of reduced size, which enables the subproblem to be solved in a new context and return the answer to the initial context. The top right shows the generation procedure to solve 1234+5678. Each separate context is indicated by a blue bubble. The arrows indicate copying of generated prompts or solutions. Bottom Left: On most problems, \method\ trains on significantly fewer tokens than the scratchpad method, saving considerable GPU memory. Bottom Right: On average, \method\ outperforms the baseline and scratchpad methods across all tasks, especially as the problems grow in size and complexity.}
    \label{fig:main_fig}
\end{figure*}

Large language models (LLM) have obtained the state-of-the-art performance on a wide set of tasks~\citep{brown2020language, taylor2022galactica, chowdhery2022palm, anil2023palm, openai2023gpt4, touvron2023llama, touvron2023llama2}. However, recent studies~\citep{anil2022exploring,dziri2023faith,zhou2023algorithms} show these models struggle to generalize to compositional tasks, where the solution depends on solutions to smaller instances of the same problem. An example task, integer addition, is shown in Figure~\ref{fig:main}. When calculating `\verb|1234 + 4567|', we first break the problem into a smaller subproblem `\verb|234 + 567|'. After obtaining the solution to this subproblem, the original problem is partially solved. Similarly, to solve `\verb|234 + 567|', we first sum `\verb|34 + 67|'. This recursion is the fundamental operation to solve compositional tasks. However, no existing approach has explicitly captured the recursive nature of compositional tasks.

In this paper, we propose a recursion-based method for LLMs to better solve compositional tasks. More specifically, we adopt a top-down approach to solve problems recursively. We train LLMs to recursively call themselves on subproblems of reduced size, recognize and solve the base case directly, and combine the solutions up the associated call stack to obtain the solution to the original problem (Figure~\ref{fig:main}). The above procedure is referred to as recursive tuning (or \method\ in short). 

The basic idea behind \method\ is motivated by two lines of work. First, recent work~\citep{nye2021work, anil2022exploring, dziri2023faith} show that training LLMs on high-quality scratchpad data, which includes intermediate steps towards solving a problem, can improve performance on certain compositional tasks such as integer addition and parity. Instead of using the intermediate steps to train models, which is computationally costly, \method\ breaks down the problems into smaller and smaller subproblems. Each subproblem runs independently within its own context in the associated call stack. The solution to each subproblem is then propagated up the call stack to produce the final solution. Since each level of the call stack only includes the information necessary to solve the current subproblem, models can more easily attend to the relevant context, improving the accuracy of solving each subproblem. Second, our tuning process is reminiscent of recent works that incorporate tool use in LLMs~\cite{schick2023toolformer, paranjape2023art}. Similar to how these models call a tool and resume generating output based on the output of the tool, with \method, the models call themselves on a subproblem and resume generating after receiving the subproblem's solution.

We empirically evaluate the performance of \method\ on three representative compositional tasks: integer addition \citep{zhou2023algorithms}, a dynamic programming problem \citep{dziri2023faith}, and the parity problem \citep{anil2022exploring, zhou2023algorithms}. Our results show \method\ improves the average performance of LLaMA 7B and LLaMA 13B on all tasks by 37.4\% and 31.9\% over baseline training. Compared to scratchpad training, our improvement is striking, with average improvements of 34.5\% and 36.7\% on LLaMA 7B and LLaMA 13B respectively. Importantly, we show \method\ saves significant GPU memory compared to the scratchpad method when training. We hope our results foster future research on recursive learning of large foundation models.
\section{Approach}

We present \method\ in this section. \method\ recursively tunes LLMs to solve compositional tasks. Specifically, the method involves (1) recursively decreasing the size of the problem, (2) solving the base case, and (3) passing the solutions up the recursion stack, solving subproblems of increasing complexity along the way.

First, with \method, an LLM recursively calls itself on subproblems of decreasing length or complexity. For example, when adding \verb|1234 + 5678|, the LLM calls itself to add \verb|234 + 678|. This call is then sent to a new context in which the LLM calls itself to add \verb|34 + 78|, which is again sent to a new context where the LLM calls itself to add \verb|4 + 8|.  

Next, the base case is solved. The base cases are easy enough to be solved directly in the same context. For the integer addition problem, the base case is to add the two least-significant digits together (e.g.,  adding \verb|4 + 8|). 

Finally, the subproblem solutions are passed up the recursive call stack. Specifically, subproblem solutions are appended directly after the associated call in the context one level up the call stack. Again, sticking with integer addition, it helps to know the sum of \verb|4 + 8| when tasked with adding \verb|34 + 78|. As such, the LLM-generated solution to \verb|4 + 8| is appended to the context tasked with solving \verb|34 + 78|. This process of propagating subproblem solutions continues up the recursive call stack until the solution to the first recursive call is passed to the initial context, which helps to solve the initial problem.

To accomplish this, we train LLMs to (1) generate recursive subproblems, (2) solve base cases, and (3) use the answers propagated up from these recursive calls in their computation for the problem in the current context. We do so by randomly generating a set of seed data from which we programmatically construct training instances for all three types (see Figure~\ref{fig:main}).

During generation, the model can designate some of its generated text to be a recursive call by enclosing the text between `\verb|Call: |' and `\verb|\n|'. Once a recursive call is made, we stop generating in the current context and prompt the model with the call in a new context. In each new context, we follow the exact same generation procedure, except for the base case where the model learns to directly output the answer rather than making another recursive call. When generation in the new context is complete, we take the subproblem solution, which is separated by the text `\verb|\nAnswer: |', and append that to the context one level higher in the associated call stack. Then we continue generation in the initial context. The pseudo-code for the recursive generation procedure is in Figure~\ref{fig:psuedocode_generation}.

\begin{figure}
    \centering
    \scriptsize
    \par\noindent\rule{\linewidth}{0.4pt}
    \par
    \begin{algorithmic}
    \Function{RecursiveGenerate}{$model, tokenizer, prompt$}
        \State $context \gets \Call{Generate}{model, tokenizer, prompt}$
        \While{$\Call{ContainsUnexecutedCall}{context}$}
            \State $call \gets \Call{ExtractCall}{context}$
            \State $result \gets \Call{RecursiveGenerate}{model, tokenizer, call}$
            \State $context \gets context + result$
            \State $context \gets \Call{Generate}{model, tokenizer, context}$
        \EndWhile
        
        \State \Return $context$
    \EndFunction
    \end{algorithmic}
    \par\noindent\rule{\linewidth}{0.4pt}
    \par
    \captionof{algorithm}{Psuedocode for the RecursiveGenerate method, a lightweight recursive wrapper around the standard generation function used with the baseline and scratchpad methods.}
    \label{fig:psuedocode_generation}
\end{figure}

In the integer addition example, the model only needs to generate one recursive call in each context. However, our method works more generally than this. Many recursive calls may be generated in a single context. For example, the dynamic programming problem we describe below requires multiple recursive calls in the initial context to solve the problem.

\section{Experiments}
We consider three tasks: integer addition, a dynamic programming problem, and the parity problem. For each task, we train 3 types of models: baseline, scratchpad, and \method. The baseline models were trained to simply output the solution to the problem. The scratchpad models were trained to generate a scratchpad~\citep{nye2021work} containing intermediate reasoning steps before generating the final solution to the problem. The \method\ models are as described above.

During evaluation, we consider both in-distribution and out-of-distribution (OOD) data. The in-distribution data are those with problem lengths that were seen in training and the OOD data are those with problem lengths longer than seen in training. For example, on the integer addition task, the training data consists of numbers with lengths up to 15 digits. Evaluation examples with 1-15 digits are considered in-distribution and examples with 16 or more digits are considered OOD.

We train LLaMA 7B and 13B~\citep{touvron2023llama} using Low-Rank Adapters~\citep{hu2022lora}. See Appendix~\ref{sec:training_details} for additional details on the training setup. Additionally, we provide results on the smaller Galactica \citep{taylor2022galactica} 125m and 1.3B parameter models, in Appendix~\ref{sec:galactica_results}.

\subsection{Experimental Setup}
We consider 3 representative compositional problems: integer addition, a dynamic programming problem, and the parity problem. Here, we describe each problem in detail, as well as how the data was constructed. Additional details are provided in Appendix~\ref{sec:data_construction} and examples are provided in Appendix~\ref{sec:example_problems}.

\paragraph{Integer addition} This problem challenges LLMs to add two integers. The input to the model is simply a prompt to add 2 numbers. For example, `\verb|45 + 97|'. Pretrained language models have some capability to perform addition without any training, but it seemingly disappears as the numbers grow in size. \citet{nye2021work} used a scratchpad to teach language models addition, and more recently~\citet{liu2023goat} taught LLaMA 7B to add numbers up to 15 digits. In both cases, there is remarkable performance degradation when adding integers larger than those seen during training. \citet{zhou2023algorithms} suggests that addition is particularly hard for LLMs since it requires precise indexing operations and non-causal propagation of the carry term. Following~\citet{liu2023goat}, we generate training data summing randomly generated integers up to 15 digits long. During evaluation, we focus on adding two numbers with the same number of digits and sample 100 problems per length up to 60 digits.

\begin{figure*}[t!]
    \centering
    \begin{subfigure}[b]{\textwidth}
        \centering
        \includegraphics[width=\textwidth]{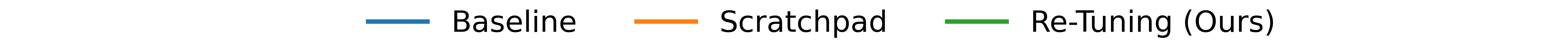}
    \end{subfigure}\\
    \vfill
    \begin{subfigure}[b]{0.32\textwidth}
        \centering
        \includegraphics[width=\textwidth]{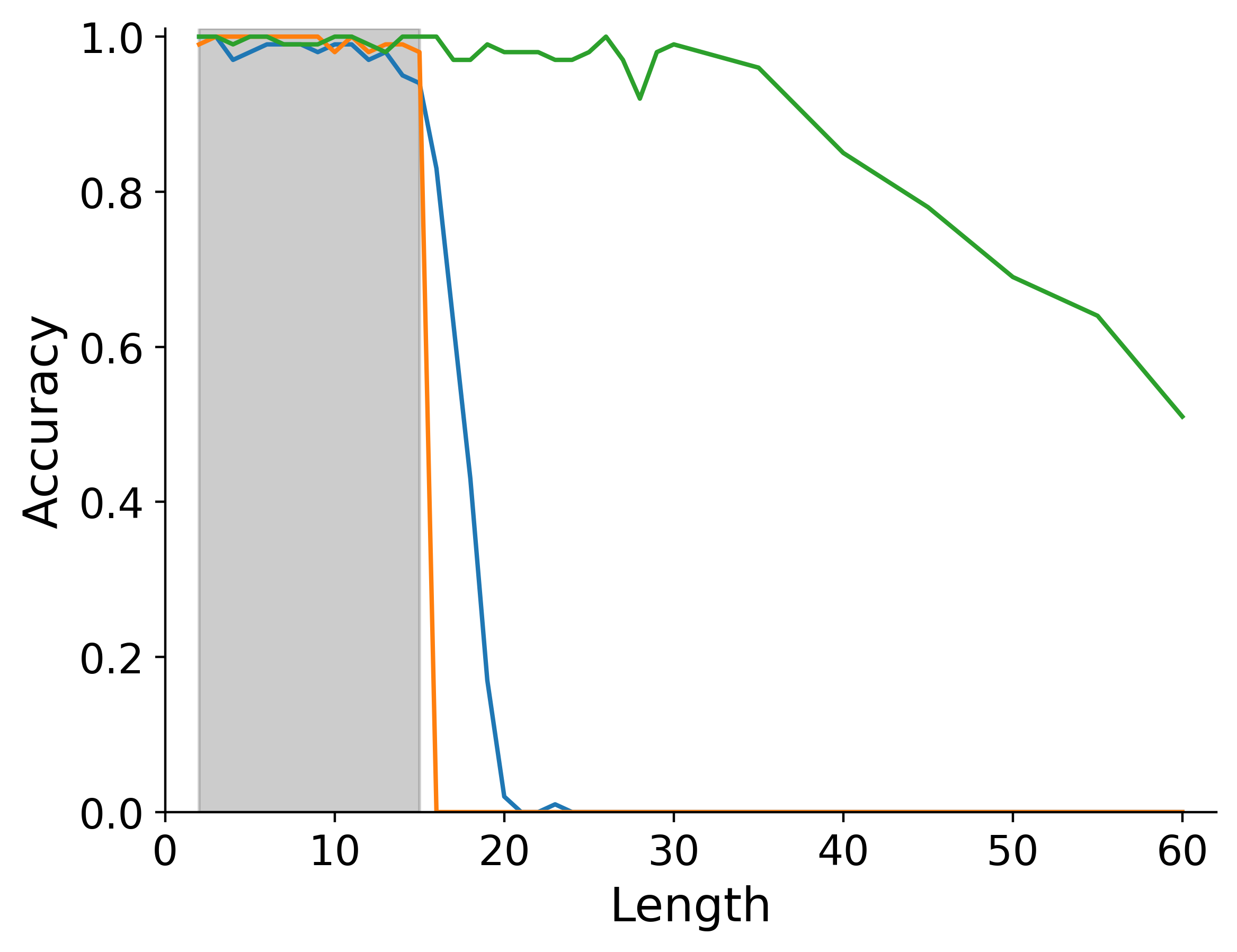}
        \caption{LLaMA 7B integer addition.}
        \label{fig:addition_7b}
    \end{subfigure}
    \hfill
    \begin{subfigure}[b]{0.32\textwidth}
        \centering
        \includegraphics[width=\textwidth]{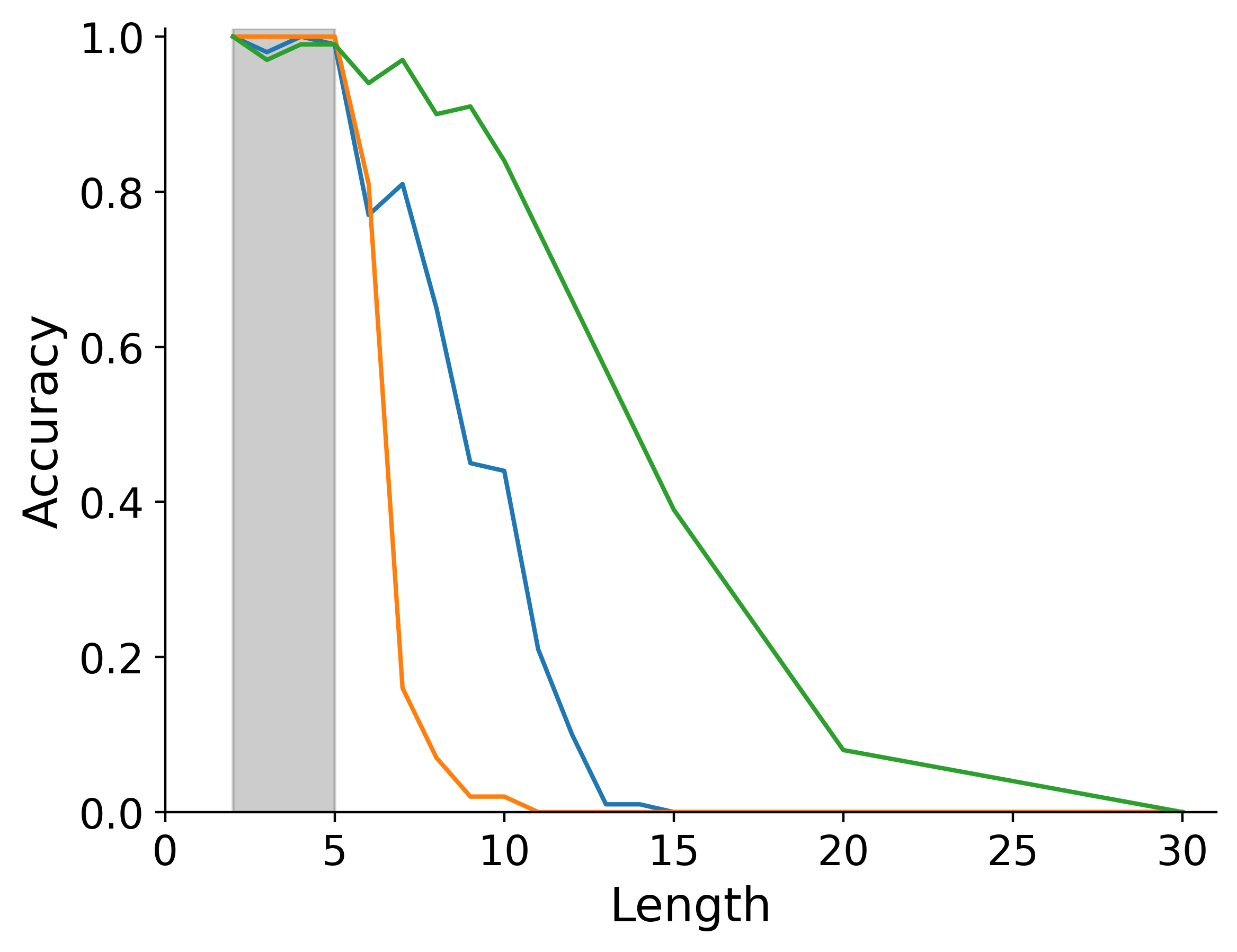}
        \caption{LLaMA 7B dynamic programming.}
        \label{fig:dp_7b}
    \end{subfigure}
    \hfill
    \begin{subfigure}[b]{0.32\textwidth}
        \centering
        \includegraphics[width=\textwidth]{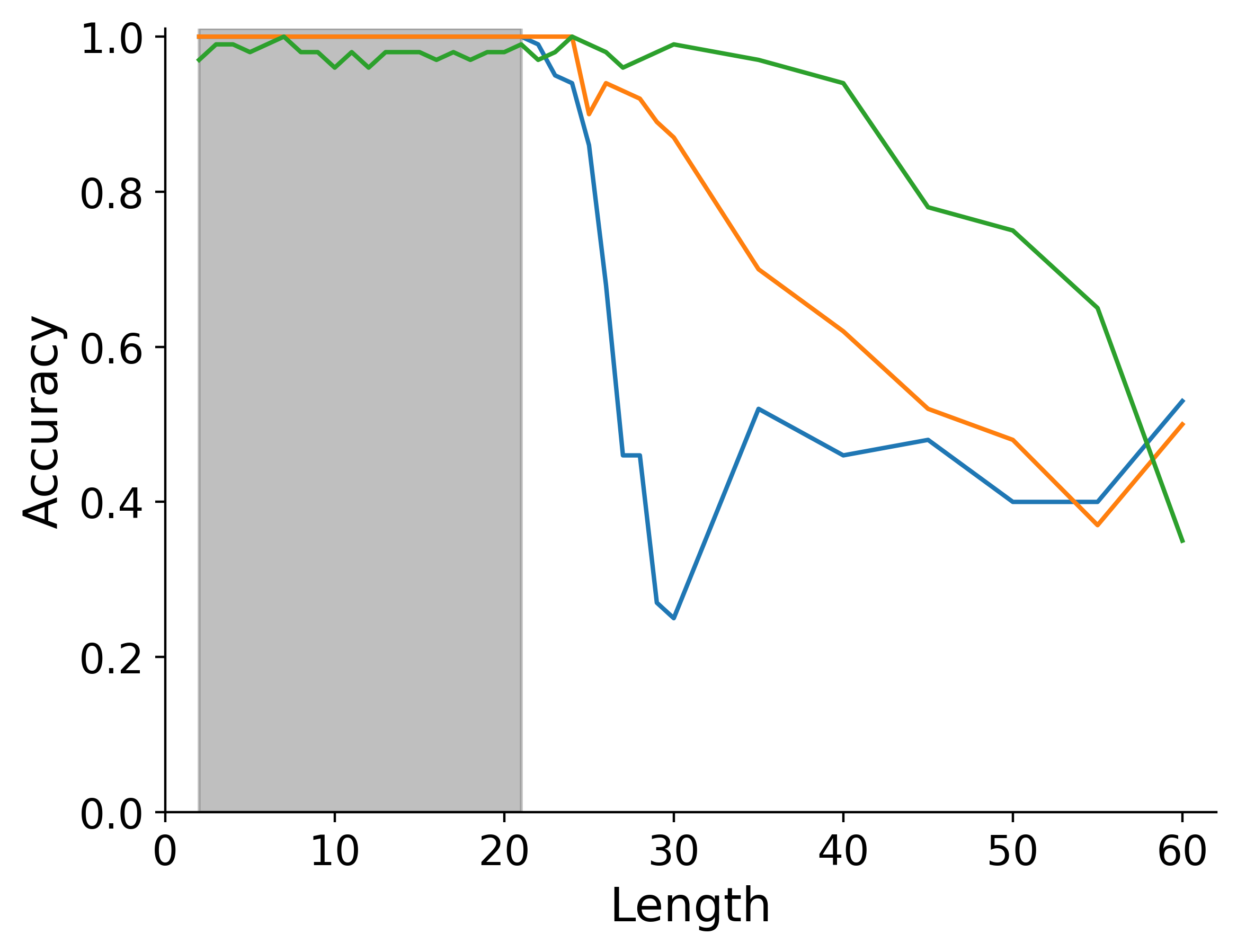}
        \caption{LLaMA 7B parity.}
        \label{fig:parity_7b}
    \end{subfigure}\\
    \par\bigskip
    \begin{subfigure}[b]{0.32\textwidth} 
        \centering
        \includegraphics[width=\textwidth]{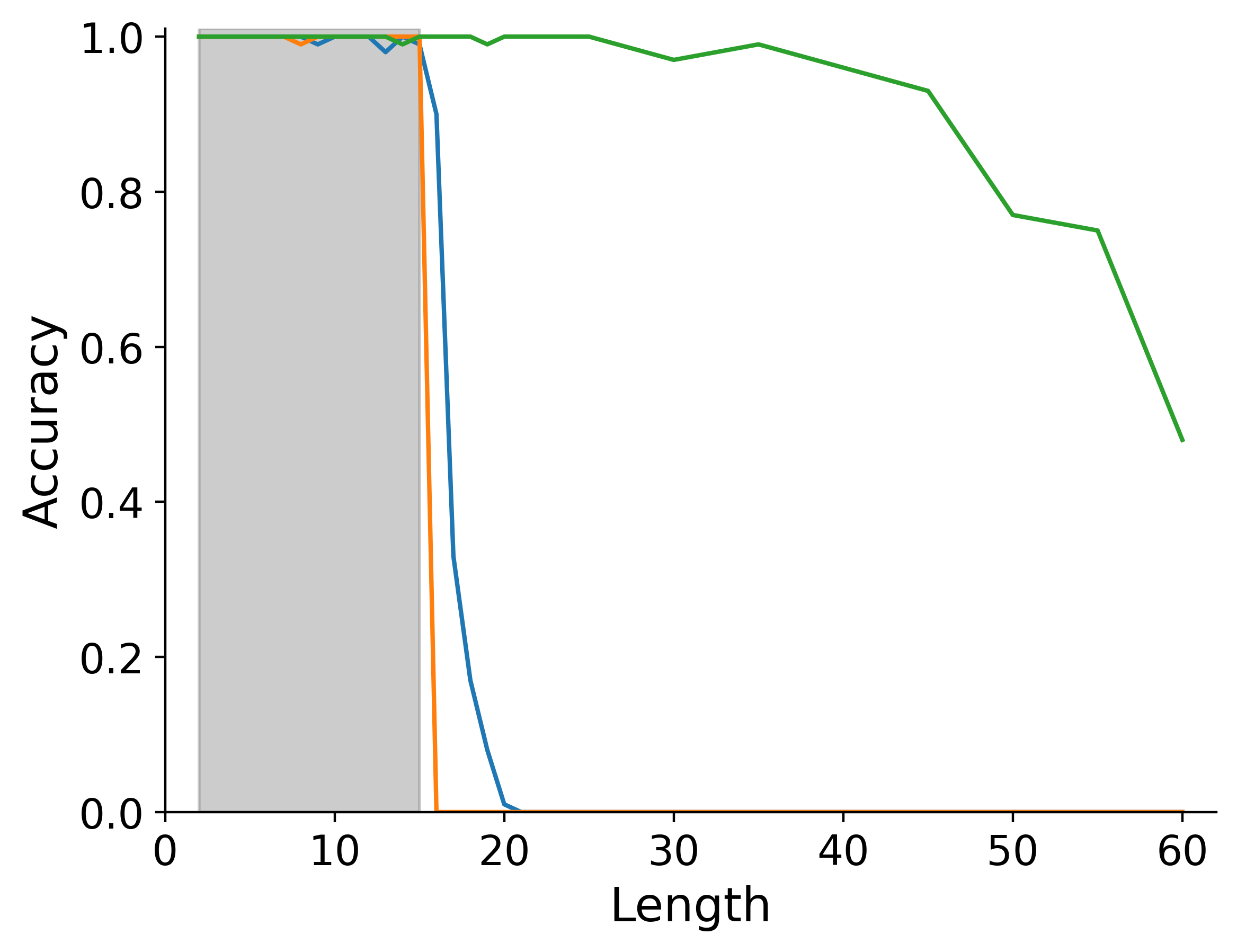}
        \caption{LLaMA 13B integer addition.}
        \label{fig:addition_13b}
    \end{subfigure}
    \hfill
    \begin{subfigure}[b]{0.32\textwidth}
        \centering
        \includegraphics[width=\textwidth]{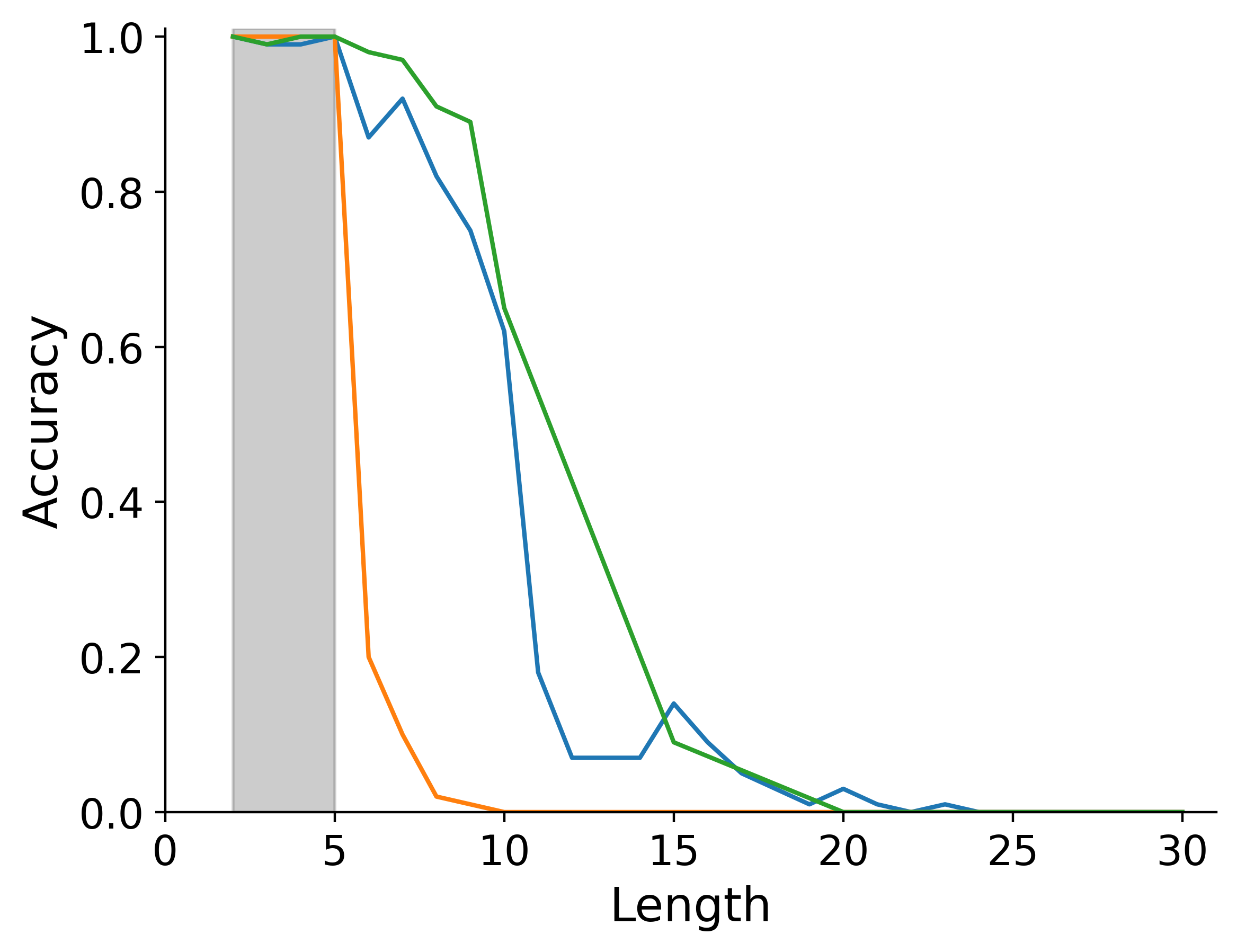}
        \caption{LLaMA 13B dynamic programming.}
        \label{fig:dp_13b}
    \end{subfigure}
    \hfill
    \begin{subfigure}[b]{0.32\textwidth}
        \centering
        \includegraphics[width=\textwidth]{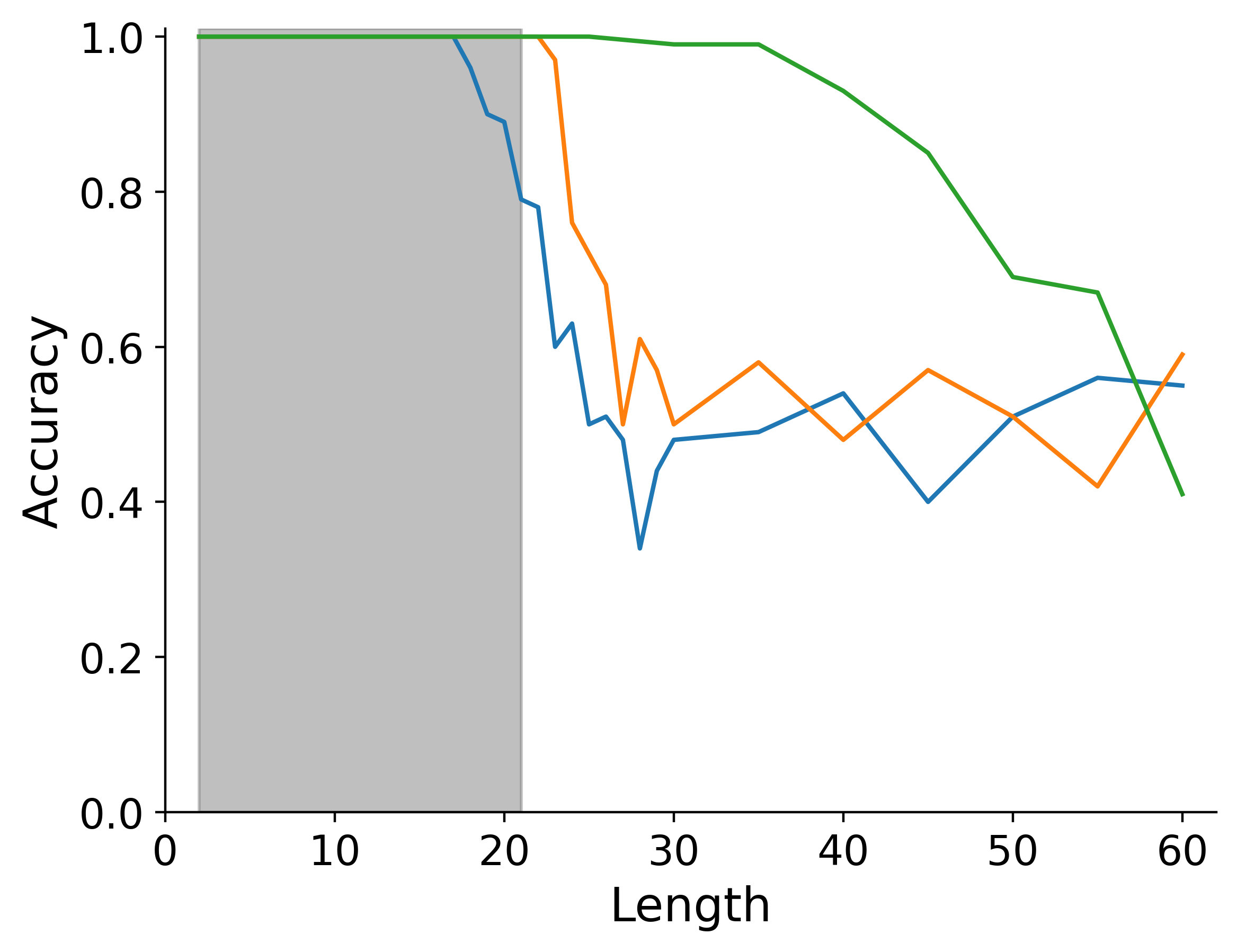}
        \caption{LLaMA 13B parity.}
        \label{fig:parity_13B}
    \end{subfigure}
    \caption{Performance of LLaMA 7B (top) and LLaMA 13B (bottom) on Addition (left), Dynamic Programming (middle), and Parity (right). The in-distribution range is shaded in gray.}
    \label{fig:main_results}
\end{figure*}

\paragraph{Dynamic programming} We borrow the dynamic programming problem recently studied by \citet{dziri2023faith}: 

\begin{quote} 
"Given a sequence of integers, find a subsequence with the highest sum, such that no two numbers in the subsequence are adjacent in the original sequence."
\end{quote}

This problem can be broken down into two steps: (1) recursively generate an array of sub-array sums and (2) recursively identify which indices correspond to the highest sum. With \method\, LLMs generate recursive calls for each of these steps, which are then solved in separate contexts. For example, consider the sequence \verb|[3, 2, -2, 5, 3]|. The subsequence with the highest sum with no adjacent numbers would contain \verb|3| (element 0) and 5 (element 3). Internally, the LLM represents the selected subsequence as a list of 1's and 2's, with 1's corresponding to numbers chosen and 2's corresponding to numbers not chosen. For the sequence \verb|[3, 2, -2, 5, 3]|, the expected output would be \verb|[1, 2, 2, 1, 2]|. Following~\citet{dziri2023faith}, we exhaustively generate all permutations of arrays up to length 5 for training, where each element is restricted to $[-5, 5]$. Evaluation is done on arrays up to length 30, again with each integer element restricted to $[-5, 5]$.

\paragraph{Parity}
The parity problem is to determine if there is an even or odd number of 1's in a binary input array. An example input is \verb|[0, 1, 0, 0]|, for which the output should be \verb|1| since the array contains an odd number of 1's. For an array with an even number of 1's, the output should be \verb|0|. This problem has been previously studied by \citet{anil2022exploring} and \citet{zhou2023algorithms}. Traditionally, this problem is solved by traversing the input array to compute the sum modulo 2, which is the method we train our models to use. We generate binary arrays up to length 21 for training. For evaluation, we sample 100 binary arrays per length up to length 60.

\subsection{Main Results}
Here we share our main results on LLaMA 7B and LLaMA 13B, across all three tasks. Results are shown in Figure~\ref{fig:main_results}, and discussed in detail in the proceeding paragraphs. Across all problems and model sizes, the \method\ method outperforms the baseline and scratchpad methods, with the clearest difference being on integer addition. In particular, \method\ exhibits significantly better OOD generalization compared to the baseline or scratchpad methods. We find this to be true even on language models with very few parameters, including Galactica 125M and Galactica 1.3B (see Appendix~\ref{sec:galactica_results}).

\paragraph{Integer addition} The \method\ method considerably outperforms the baseline and scratchpad methods. The scratchpad method performs the worst, achieving 0\% accuracy on every problem longer than those seen during training on both LLaMA 7B and LLaMA 13B. The baseline method has non-zero OOD accuracy for problems up to length 20, but accuracy falls to 0\% on longer problems with both models. In contrast, the \method\ method maintains near-perfect accuracy in regimes where the baseline and scratchpad models have 0\% accuracy, only falling below 90\% accuracy on problems of length 40 and 45 for LLaMA 7B and LLaMA 13B respectively. Astonishingly, with \method, both models maintain near 50\% accuracy on adding up to 60 digit numbers. The model by~\citet{liu2023goat}, which is also trained on addition up to 15 digits, has similar OOD performance to our baseline models, and falls to 0\% accuracy when adding 21-digit numbers.

\paragraph{Dynamic programming} Again, \method\ outperforms both the baseline and scratchpad approaches, though the gap between \method\ and baseline is narrower for LLaMA 13B than it is for LLaMA 7B. Still, with \method, both models achieve near 90\% accuracy on problems of length 10, twice as long as the longest examples in the training data. Moreover, on problems of length 15, LLaMA 7B achieves 40\% accuracy with \method\ and 0\% accuracy with the baseline and scratchpad methods. \citet{dziri2023faith} trains and evaluates GPT3 models with and without scratchpad. Both reach 0\% accuracy on problems of length 10, which is similar to our scratchpad results but slightly worse than our baseline results.

\paragraph{Parity} \method\ performs as well or better on inputs smaller than size 60. Specifically, the accuracy of the baseline and scratchpad methods falls to that of random chance (50\%) on problems with lengths greater than 40 on both models. Meanwhile, with \method, LLaMA 7B and LLaMA 13B maintain an accuracy over 90\% on problems with the length 40.
\section{Analysis and Further Discussion}
In this section, we conduct additional experiments in order to better understand the effects of various mechanisms behind \method.

\subsection{Ablation Study}
For all tasks, as the problem size grows, so does the number of unique possible problems. For example, there are more combinations of 10-digit addition problems than there are 2-digit addition problems. If we randomly sample problems from the space of all possible problems up to some length, then the distribution of problems will be skewed toward longer problem instances. Due to \method's recursive design, it's important that an appropriate number of small problems are included in the training data. As such, we upsample examples with shorter lengths and downsample examples with longer lengths. Our resampling methodology is described in Appendix~\ref{sec:data_construction} 

To better understand the impact of resampling, we train LLaMA 7B using integer addition data with and without resampling. Both models saw the same number of training examples. We collect results on 100 problems of length 5, 20, 35, and 50. Results are shown in Table~\ref{table:ablation}.

\begin{table}
    \centering
    \small
    \begin{tabular}{lcccc}
        \toprule
         & 5 & 20 & 35 &50 \\ \midrule
        w/ Resampling & 1.0 & 0.98 & 0.96 & 0.69\\
        w/o Resampling & 1.0 & 0.97 & 0.73 & 0.0\\
    \bottomrule
    \end{tabular} 
    \caption{Ablation over resampling approach during \method\ training.}
    \label{table:ablation}    
\end{table}
\begin{figure*}[ht]
\centering
\includegraphics[width=0.9\textwidth]{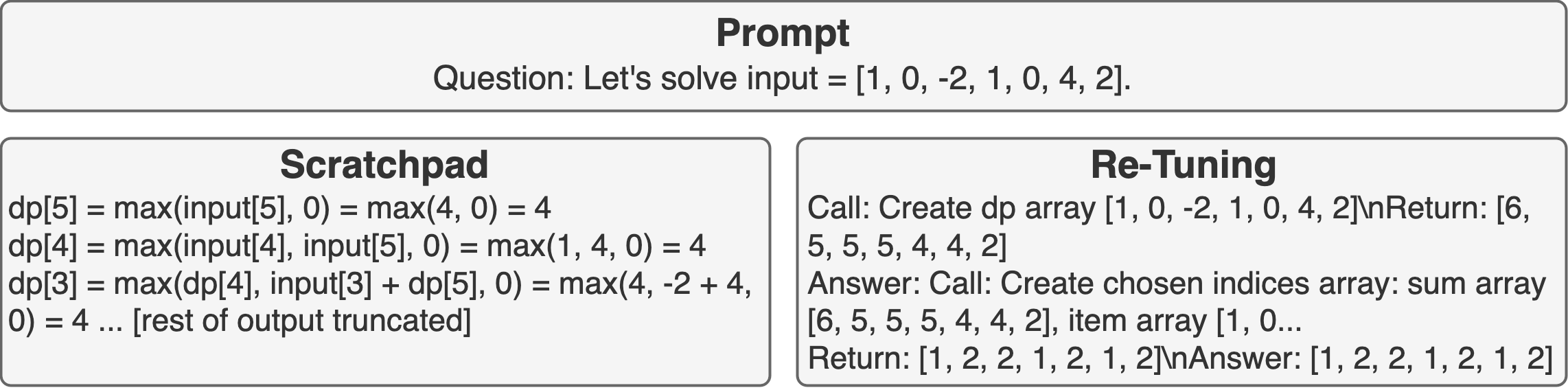}
\caption{Case study on dynamic programming problem. With scratchpad, the model makes an indexing error, while with \method, the model correctly generates the recursive call.}
\label{fig:case_study_dp_scratchpad}
\end{figure*}

While both trained models perform well on problems up to length 20, the superiority of the resampling approach becomes clear on longer problems. At a problem length of 35, the resampling model achieves 96\% accuracy, while the accuracy of the model trained without resampling is only 73\%. At a problem length of 50, the model trained without resampling fails to correctly solve a single problem instance. However, the model trained with resampling maintains an accuracy of 69\%, which suggests that resampling is an important contributor towards the success of \method. 

\subsection{Case Study}
The scratchpad models often make errors with indexing operations. For example, on the dynamic programming problem, the training data includes arrays up to size 5. In Figure~\ref{fig:case_study_dp_scratchpad} we see that the model-generated scratchpad indexes element 5 instead of element 6 of the array of sub-array sums (dp array), which is incorrect on an input array of length 7. Once the model makes this indexing error on the scratchpad it is unable to recover. In other cases, the scratchpad method correctly generates the text for "dp[6]" but fails to populate the subsequent expressions with the correct values from the input array. In contrast, the \method\ method is shown in Figure~\ref{fig:case_study_dp_scratchpad}. Only the initial context is shown to save space. With \method, the model is able to generate recursive calls correctly with no difficulty indexing, enabling it to correctly solve the problem.

\subsection{Error Analysis}
\label{sec:error_analysis}
In order to better understand the types of errors made by \method\ models, we perform extensive error analysis on each task. For each task, we randomly sample 20 problems per problem length and use LLaMA 7B with \method\ to generate the outputs. We categorize these samples into the following error types: 

\begin{itemize} 
\item \textbf{Call error}: At some point in the call stack, an incorrect recursive call is made. As a result, the input prompt to the new context is incorrect. 
\item \textbf{Compute error}: This error can manifest either because the base case is incorrectly solved, or at some point in the call stack the model returns the wrong solution to a subproblem even though the correct answer to its recursive call was received. As a result, the answer returned by the current context to the earlier context will be incorrect.
\item \textbf{Restoration error}: A restoration error occurs if, at some point in the call stack, a call error or compute error is made, yet later recovered such that the final answer to the prompt in the initial context is correct. Importantly, since the model is able to recover, instances of restoration errors are classified as correct.
\item \textbf{No error}: In order for a problem instance to be free of errors, each recursive call must be correct, the base case must be solved correctly, and the correct answers are propagated up the call stack, leading to the correct final answer.
\end{itemize}

\begin{figure*}[t!]
    \centering
    \begin{subfigure}[b]{\textwidth}
        \centering
        \includegraphics[width=\textwidth]{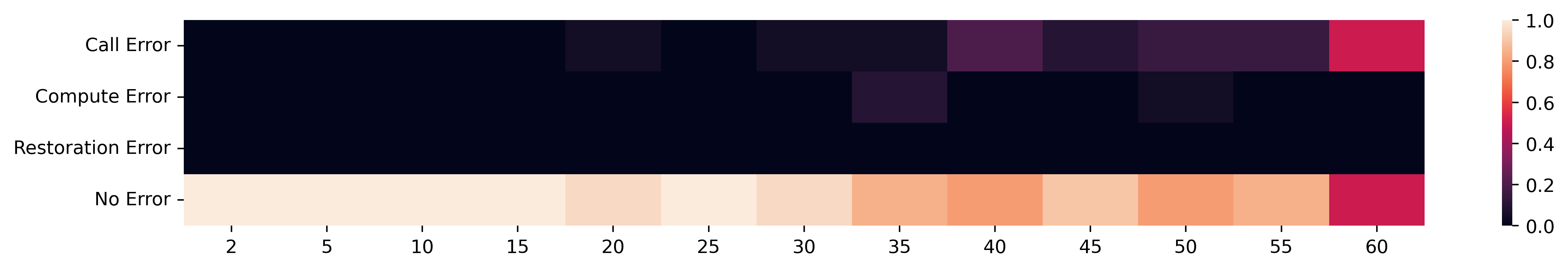}
        \caption{Errors on integer addition.}
        \label{fig:errors_addition}
    \end{subfigure}\\
    \par\bigskip
        \begin{subfigure}[b]{0.48\textwidth}
        \centering
        \includegraphics[width=\textwidth]{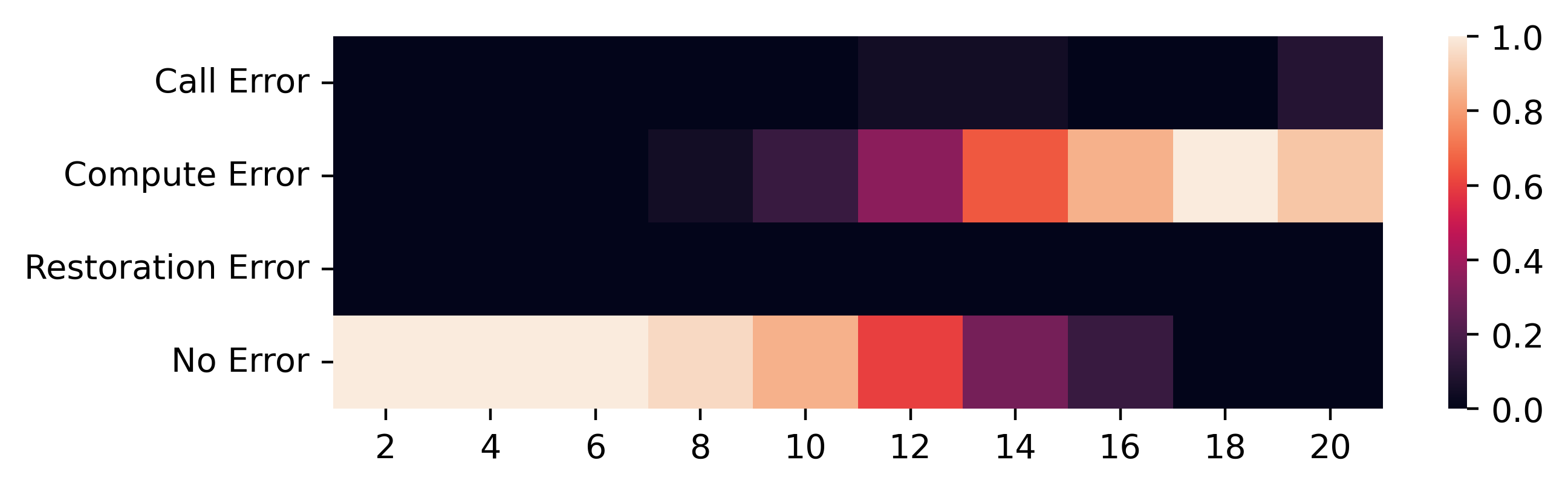}
        \caption{Errors on dynamic programming (subproblem 1).}
        \label{fig:errors_dp1}
    \end{subfigure}
    \hfill 
    \begin{subfigure}[b]{0.48\textwidth}
        \centering
        \includegraphics[width=\textwidth]{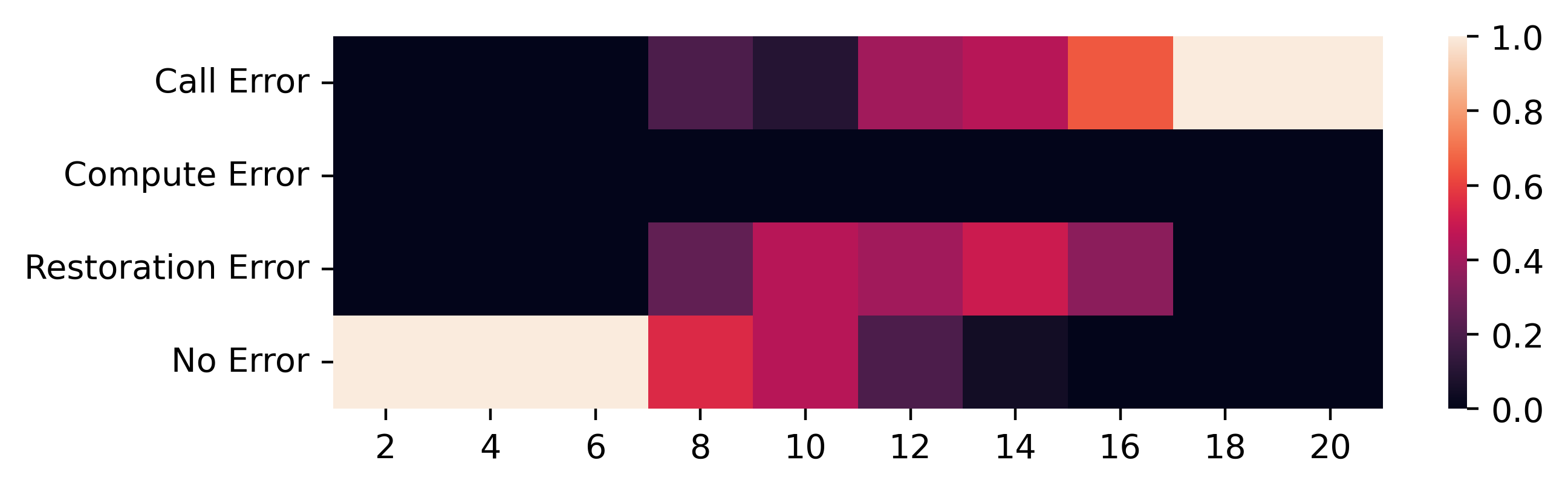}
        \caption{Errors on dynamic programming (subproblem 2).}
        \label{fig:errors_dp2}
    \end{subfigure}\\
    \par\bigskip
    \begin{subfigure}[b]{\textwidth}
        \centering
        \includegraphics[width=\textwidth]{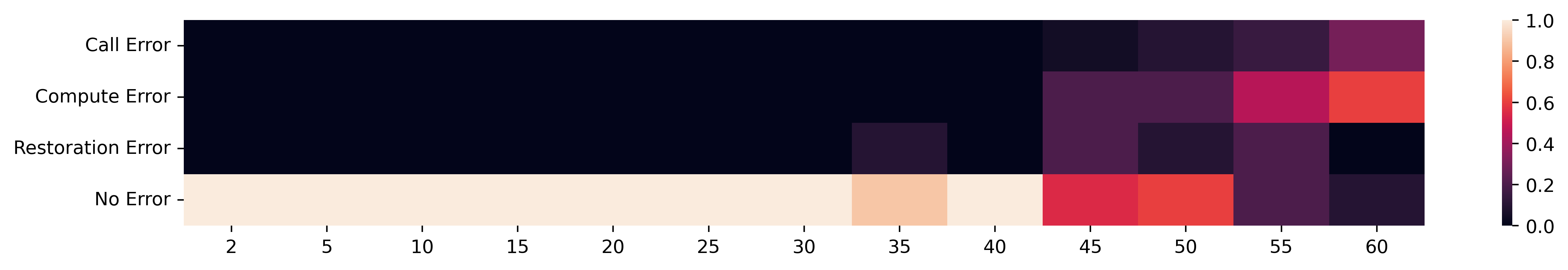}
        \caption{Errors on parity.}
        \label{fig:errors_parity}
    \end{subfigure}
    \caption{Error classifications for each problem across samples of 20 instances per problem lengths.}
    \label{fig:error_analysis}
\end{figure*}

Figure~\ref{fig:error_analysis} displays the error classifications for each problem on LLaMA 7B. Importantly, across all problems, the prevalence of errors increases with the size of the problem.

\paragraph{Integer addition} On the integer addition task, very few call errors and compute errors occur before a problem size of 30. On more complex instances call errors frequently occur, suggesting that the model has a difficult time constructing the subproblem. Importantly, this aligns with \citet{zhou2023algorithms}, which suggests that simply copying long strings of text with repeating characters is a difficult task for transformer-based models to perform. Furthermore, the lack of restoration errors suggests that once a call error is made, the model has a very hard time recovering.

\paragraph{Dynamic programming} For the dynamic programming problem, we perform error analysis on each subproblem separately. The first subproblem deals with constructing the array of sub-array sums, while the second subproblem identifies which indices correspond to the maximum sum. While compute errors occur most frequently on the first subproblem, call and restoration errors occur more frequently on the second subproblem. This checks out, as the first subproblem requires a rather simple call, but involves a more complex step to compute the answer, whereas the second subproblem contains a more involved recursive call, but an easier computation to produce the index array. Furthermore, the prevalence of restoration errors on subproblem 2 suggests that these call errors are easier to recover from than the computer errors made in subproblem 1. 

\paragraph{Parity} For the parity problem, we again see very few errors of any type before a problem size of 40. In contrast with the addition problem, the majority of errors made on the parity problem are compute errors, not call errors. This is rather unintuitive, as the addition operation is seemingly much harder than the possible parity flip in the parity problem.

\subsection{Improved Sample Efficiency}
\label{sec:sample_efficiency}
\begin{figure*}[t!]
    \centering
    \begin{subfigure}[b]{\textwidth}
        \centering
        \includegraphics[width=\textwidth]{figures/src/legend_7b.png}
    \end{subfigure}\\
    \vfill
    \begin{subfigure}[b]{0.32\textwidth}
        \centering
        \includegraphics[width=\textwidth]{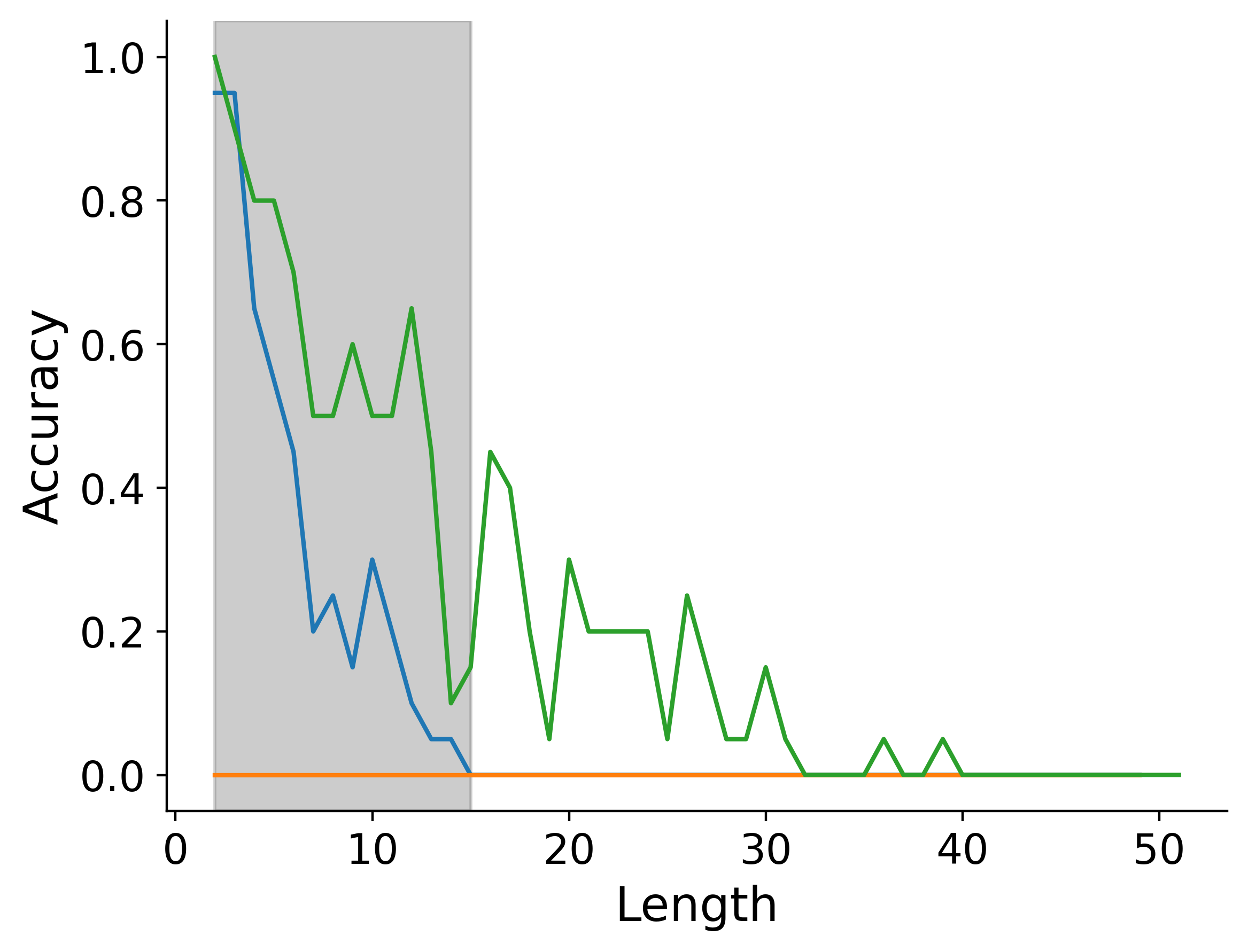}
        \caption{10 examples per length.}
        \label{fig:low_data_10}
    \end{subfigure}
    \hfill
    \begin{subfigure}[b]{0.32\textwidth}
        \centering
        \includegraphics[width=\textwidth]{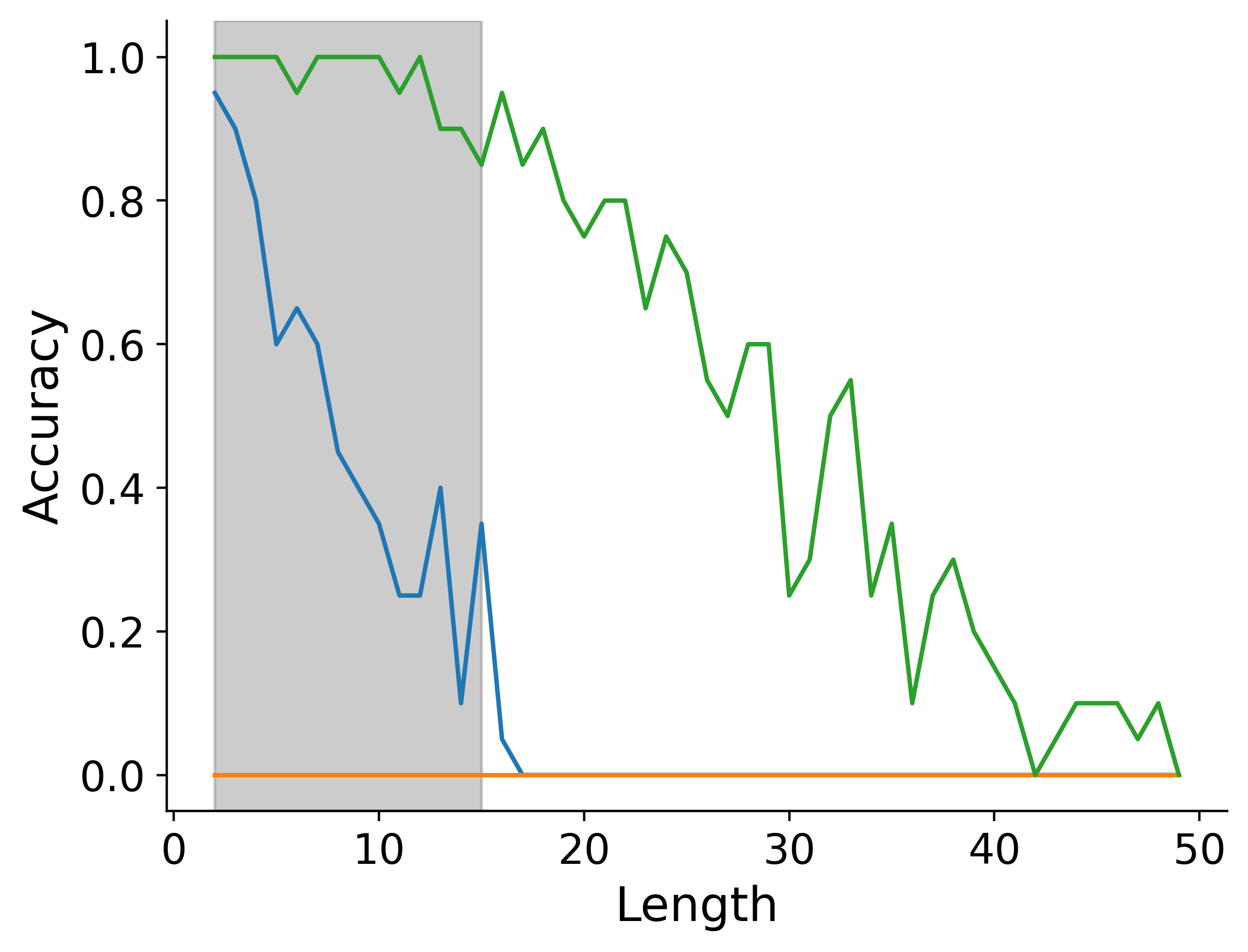}
        \caption{25 examples per length.}
        \label{fig:low_data_25}
    \end{subfigure}
    \hfill 
    \begin{subfigure}[b]{0.32\textwidth}
        \centering
        \includegraphics[width=\textwidth]{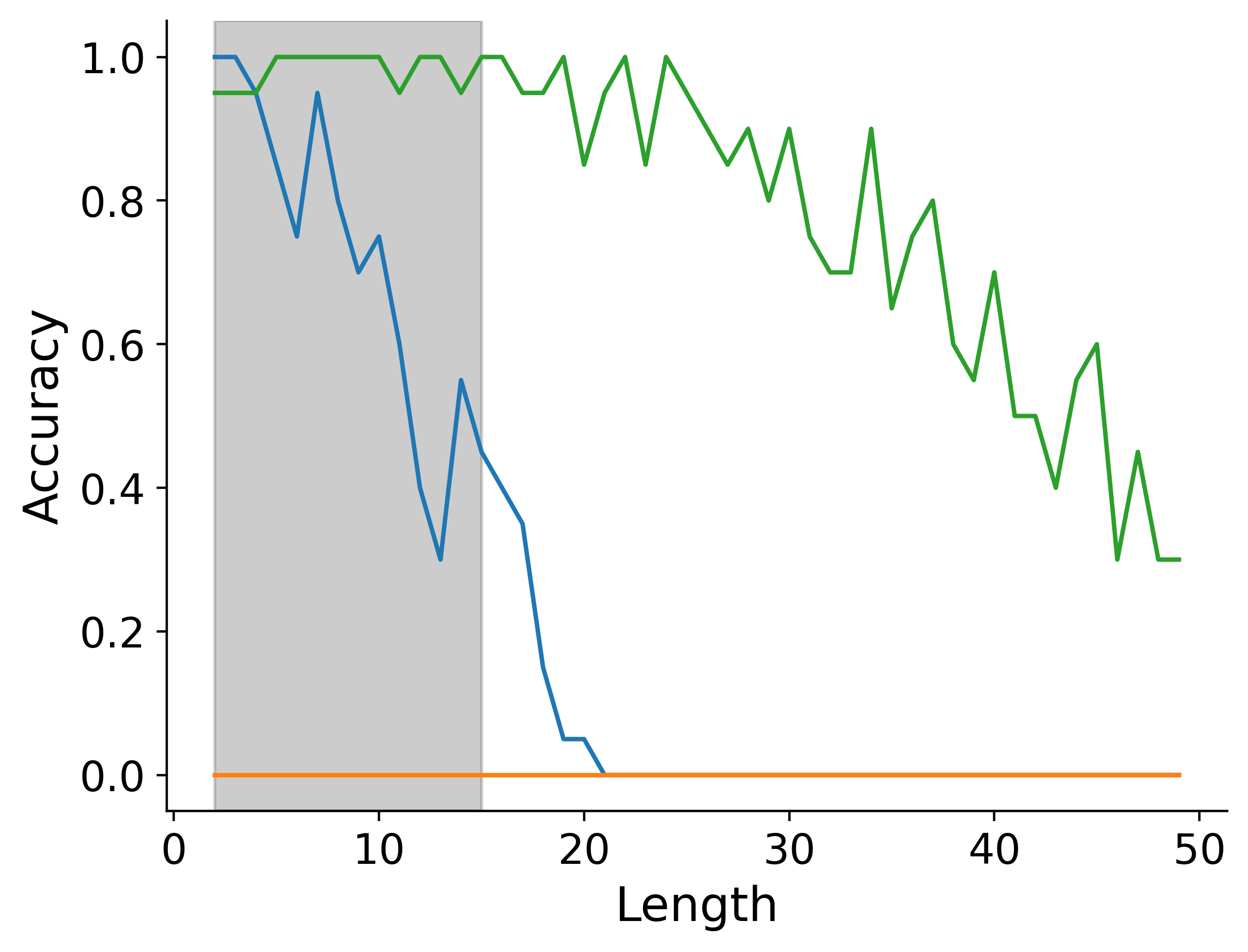}
        \caption{50 examples per length.}
        \label{fig:low_data_50}
    \end{subfigure}
    \caption{Results on integer addition with limited training data of 10 examples per length (left), 25 examples per length (middle), and 50 examples per length (right) on LLaMA 7B. Note that the scratchpad model does not get any problems correct.}
    \label{fig:limited_data_results}
\end{figure*}

We also run experiments to see the performance of \method\ in the low-data regime on integer addition. The results are in Figure~\ref{fig:limited_data_results}. For each experiment, we construct training data consisting of $n$ examples for each problem length, where $n$ is 10, 25, and 50 and the numbers contain between 1 and 15 digits. For example, when $n$ is 10, the model will see 10 examples of adding two 1-digit numbers, 10 examples of adding two 2-digit numbers, etc. So, it sees 150 examples in total when $n$ is 10. We train baseline, scratchpad, and \method\ variants of LLaMA 7B on these examples for 5 epochs each. Unlike the other experiments, we do not use any resampling here.

After seeing only 50 examples per problem length, the \method\ model achieves performance close to the \method\ model in the full-data regime above. In contrast, the baseline model has much worse performance than the baseline model in the full-data regime. With only seeing 10 examples per problem length in training, the \method\ model is comparable to the baseline model that sees 50 examples per problem length in training, a 5x efficiency increase.

\subsection{Prompt Sensitivity}
Here, we explore the sensitivity of \method\ models to various prompts during inference. Specifically, we take our best LLaMA 7B checkpoint trained on the integer addition task with \method\ using the prompt `\verb|{num_1} + {num_2}\nSolution: |', and we evaluate the model using several alternative prompt formats for inference. Results are shown in Table~\ref{table:prompt_sensitivity} on 100 problems of length 5, 20, 35, and 50.

\begin{table}
    \centering
    \scriptsize
    \begin{tabular}{lcccc}
        \toprule
         & 5 & 20 & 35 &50 \\ \midrule
        ``\verb|{num_1} + {num_2}\nSolution: |'' & 1.0 & 0.98 & 0.96 & 0.69\\
        ``\verb|{num_1} + {num_2}\nAnswer: |'' & 1.0 & 1.0 & 0.86 & 0.65\\
        ``\verb|{num_1} + {num_2}\n |'' & 1.0 & 0.98 & 0.88 & 0.67\\
        ``\verb|{num_1} - {num_2}\nSolution: |'' & 0.28 & 0.26 & 0.18 & 0.07\\
    \bottomrule
    \end{tabular} 
    \caption{Prompt sensitivity analysis on LLaMA 7B with \method\ on the integer addition problem.}
    \label{table:prompt_sensitivity}    
\end{table}

The results suggest that during inference, \method\ is not very sensitive to minor deviations in the prompt. The first prompt in Table~\ref{table:prompt_sensitivity} is the prompt used during training. The second and third prompts include small prompt deviations and result in slightly worse performance on longer problems. Specifically, the 2nd prompt uses `\verb|Answer: |' in an attempt to have the model skip the recursive call, but it appears that \method\ is robust against such attacks. \method\, however, is not robust against the fourth prompt, which adversarially prompts for subtraction rather than addition, resulting in significantly worse performance. 

\subsection{Why is \method\ so Effective?}
As discussed in Section~\ref{sec:sample_efficiency}, \method\ exhibits significantly higher sample efficiency than the scratchpad or baseline methods. However, there are other factors at play that also contribute to the success of \method. 

With \method, an LLM generates a recursive call with a subproblem to be solved in a separate recursive context. Once the subproblem is solved, the solution is returned to the original context. This approach has two advantages. First, the computation required in each context is limited. In contrast, the scratchpad approach requires a long chain of computations must be performed in the same context. Second, each context includes only the necessary information to solve the current subproblem, as information that is irrelevant to solving the subproblem is filtered out. For example, to add 2 10-digit numbers, one of the prompts generated by \method\ will be to add 2 3-digit numbers. This prompt has filtered out 7 digits from each number that would be irrelevant when adding these 3-digit numbers. This is in contrast to scratchpad prompting, where the model would also add these 3-digit numbers, but would have the full 10-digit numbers in context.

Furthermore, an important contributor to the success of LLMs on arithmetic tasks is the consistent tokenization of numbers~\citep{nogueira2021investigating, kim-etal-2021-seen}. Fortunately, the LLaMA family of models (as well as the Galactica models presented in Appendix~\ref{sec:galactica_results}) performs digit-level tokenization, where long numbers are split into individual digits for tokenization. Since all of our tasks are arithmetic in nature, it's likely that some of the success of \method\ can be attributed to digit-level tokenization.

\subsection{Efficiency Comparison}
Due to \method's high degree of sample-efficiency (see Figure~\ref{fig:limited_data_results}) and shorted training sequences relative to the scratchpad method (see Figure~\ref{fig:context}), \method\ is an efficient and effective training paradigm to improve the performance of LLMs on compositional tasks. Still, generation takes longer with \method\ than with baseline or scratchpad methods. This is because \method\ generates additional tokens related to calling the subproblem(s) within each context. To better understand this, we track the average generation time (in seconds) on the integer addition task across a selection of problem lengths. Specifically, we use LLaMA 13B at half-precision running on a single NVIDIA A100 GPU. Table~\ref{table:generation_times} displays the results.

\begin{table}
    \small
    \centering
        \begin{tabular}{lccc}
        \toprule
        Length & \method\ & Scratchpad & Baseline \\ \midrule
        5 & 4.462 & 2.492 & 0.346\\
        30 & 76.494 & 32.509 & 1.496 \\
        60 & 265.002 & 80.699 & 2.030 \\
        \bottomrule
    \end{tabular}
    \caption{Generation times (in seconds) on the integer addition task with LLaMA 13B for a selection of problem lengths.}
    \label{table:generation_times}
\end{table}

Though this would appear to be a limitation with \method, there are two additional factors to consider. First, \method\ prioritizes effectiveness over efficiency. Though all three methods see near-perfect accuracy on problems of length 5, the baseline and scratchpad approach fail to solve a single problem of length 30 or 60 correctly, while \method\ maintains near-perfect accuracy on problems of length 30, and near 50\% accuracy on problems of length 60. Second, unlike the baseline and scratchpad methods, \method\ can leverage cache-based optimizations to retrieve solutions to subproblems without the need to generate using the model, saving time and compute resources.
\section{Related Work}
Several works have explored the length generalization ability of LLMs on compositional problems.~\citet{dziri2023faith} suggests that LLMs solve compositional tasks via ``linearized subgraph matching'' and thus fail to learn the underlying algorithm necessary to solve more complex problem instances. \citet{anil2022exploring} showed that training on a combination of in-context learning and scratchpad prompting could enable better performance. Similarly, \method\ involves training pretrained LLMs to make recursive calls in order to improve performance on compositional tasks. Other works have studied length generalization on small, purpose-built transformer models. \citet{lee2023teaching} and \citet{zhou2023algorithms} showed that training small transformer models from scratch on scratchpad data could enable better length generalization. Recently, \citet{mcleish2024transformers} showed that transformers can achieve strong OOD performance on addition by using special positional embeddings.

Various papers have explored the idea of LLMs prompting themselves or other LLMs, although, to our knowledge, no papers explicitly train a language model to do so. \citet{zhou2023leasttomost} prompts a language model to break a problem down into simpler steps and then prompts itself to solve each step individually in a sequential, non-recursive manner. Similar methods have been proposed as a way to improve the logical consistency of the generated responses~\citep{crispino2024agent, imani2023mathprompter, madaan2023selfrefine}. \citet{weston20232} use a language model to generate prompts by extracting relevant information from the context. Similarly, \method\ generates a recursive call that includes the relevant information for solving a simpler subproblem.

A recent topic of interest has been teaching language models to use tools~\citep{hsieh2023tool, parisi2022talm, schick2023toolformer, qin2023toolllm, paranjape2023art, mialon2023augmented}, which often involve stopping the generation and waiting for the tool output before continuing with generation. \method\ can be interpreted as teaching LLMs to use themselves as a tool.

Several papers have investigated the ability of language models on arithmetic tasks~\citep{shen2024measuring, lee2023teaching, liu2023goat, nye2021work, nogueira2021investigating, kim-etal-2021-seen, duan2023interpolation}. In many cases, it was noticed that performance was significantly worse on problems longer than those seen during training.
\section{Conclusion}
We study the problem of solving compositional tasks with large language models. We propose a new tuning paradigm that decomposes the original compositional problem into smaller and smaller instances of the same type, solves each, and combines the results to produce the final answer. To the best of our knowledge, our method is the first to utilize the recursive property of compositional tasks. Experimental results on three representative compositional tasks demonstrate the effectiveness of our method. Our method not only significantly outperforms standard training and state-of-the-art methods, especially on out-of-distribution problem instances, but is also more memory efficient during training. We hope our method can be applied to more tasks where recursive computation is inherent and computational resources are limited.
\section*{Limitations}
\method\ has shown better accuracy and better sample efficiency than standard methods. However, it does have some disadvantages. \method\ takes longer to generate responses than standard prompting because it generates recursive calls in addition to generating the final answer. The inference procedure for \method\ is also more complex than standard inference since we need to extract text from contexts, check if there is a generated prompt in the text, and recursively generate using the generated prompts.

\bibliography{custom}
\appendix
\section{Experimental Details}
In this section, we provide additional experimental details, including details related to the synthetic construction of the data, training process, and inference pipeline. 

\subsection{Data Construction}
We highlight our pipeline for synthetic data construction in this section. 
\label{sec:data_construction}

\paragraph{Seed data} To construct the training data, we start by randomly generating a collection of seed data. On the dynamic programming and parity problems, this seed data is exhaustive (e.g., all possible binary arrays up to length 20). On the integer addition task, we randomly generate 304,000 pairs of numbers up to 15 digits long. Next, we generate the recursive solution, including the solutions to the recursive sub-problems, for each instance of the seed data. The union of the seed data and recursive sub-problems from the training data, which we format according to the method (baseline, scratchpad, and \method). 

\begin{figure*}[t!]
    \begin{subfigure}[t]{0.32\textwidth}
        \centering
        \includegraphics[width=\textwidth]{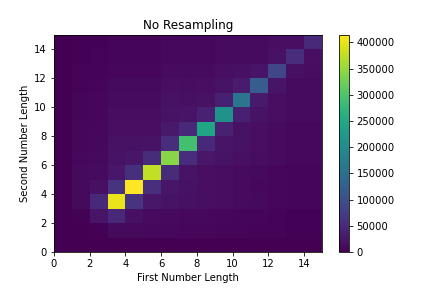}
        \caption{Integer addition (before resampling).}
    \end{subfigure}
    \hfill
    \begin{subfigure}[t]{0.32\textwidth}
        \centering
        \includegraphics[width=\textwidth]{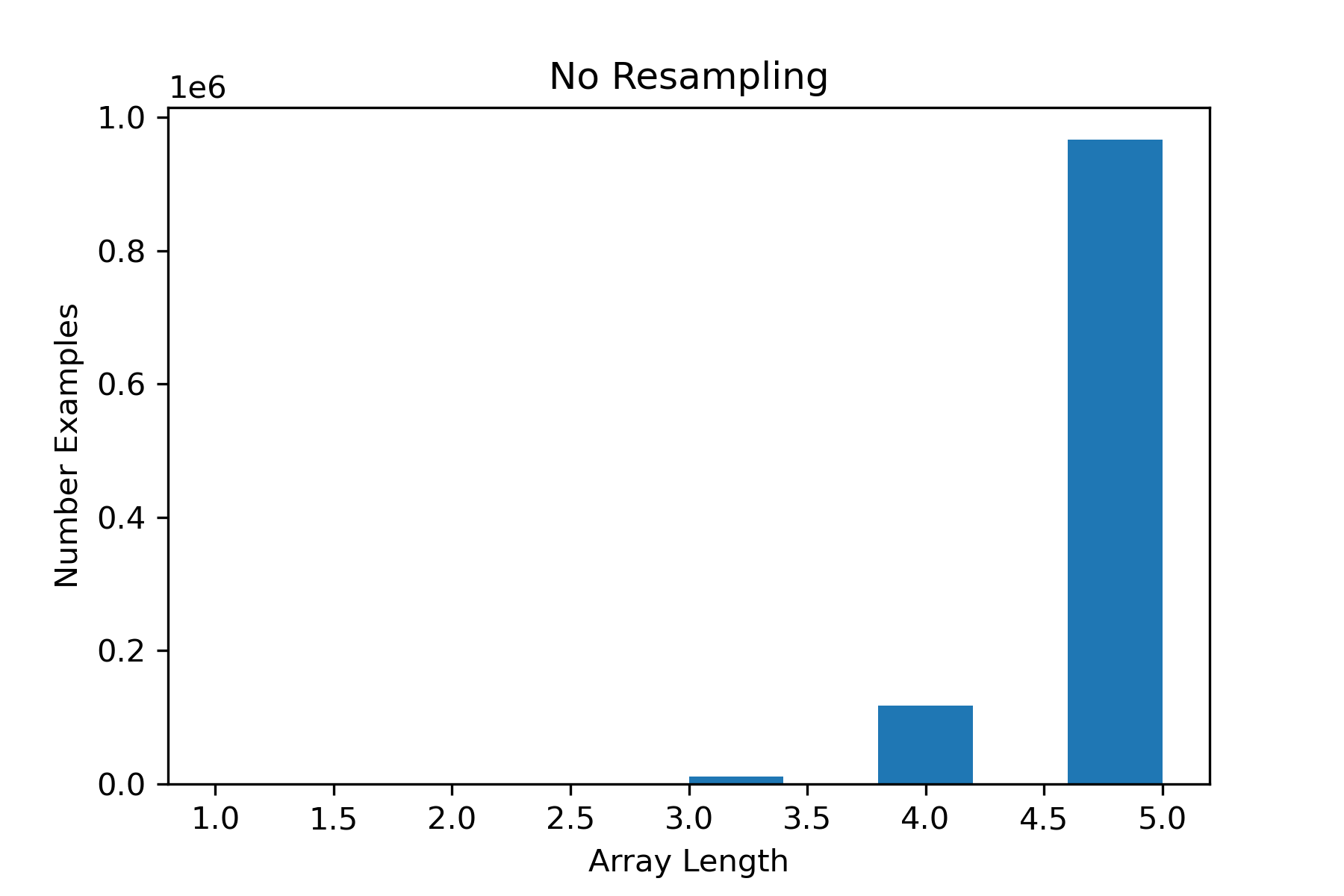}
        \caption{Dynamic programming (before resampling).}
    \end{subfigure}
    \hfill
    \begin{subfigure}[t]{0.32\textwidth}
        \centering
        \includegraphics[width=\textwidth]{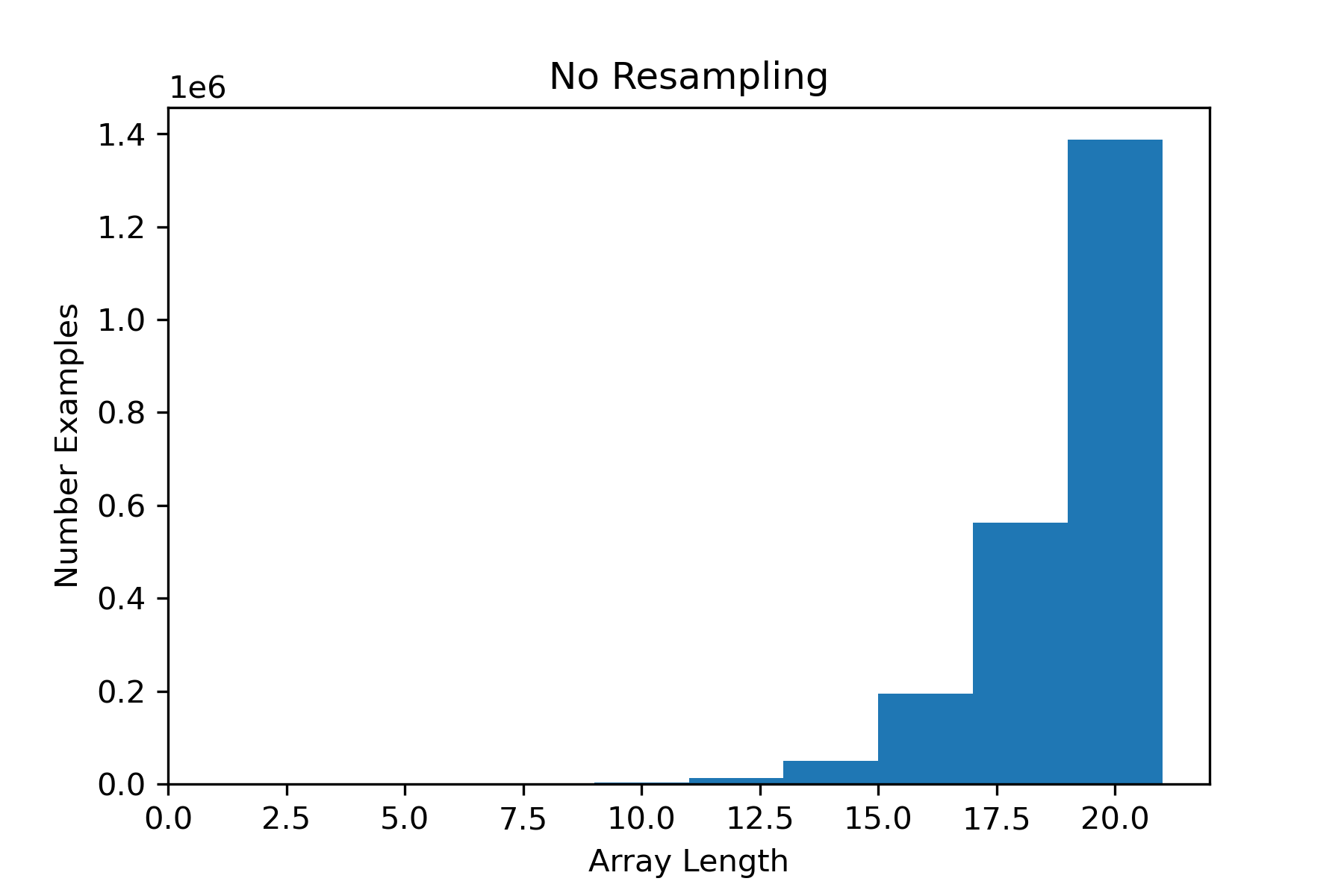}
        \caption{Parity (before resampling).}
    \end{subfigure}\\
    \par\bigskip
    \begin{subfigure}[t]{0.32\textwidth} 
        \centering
        \includegraphics[width=\textwidth]{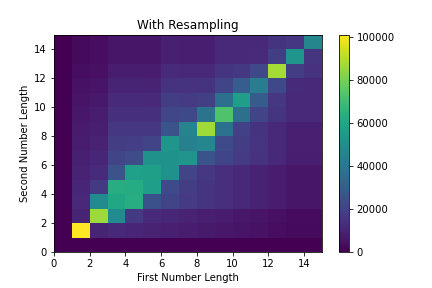}
        \caption{Integer addition (after resampling).}
    \end{subfigure}
    \hfill
    \begin{subfigure}[t]{0.32\textwidth}
        \centering
        \includegraphics[width=\textwidth]{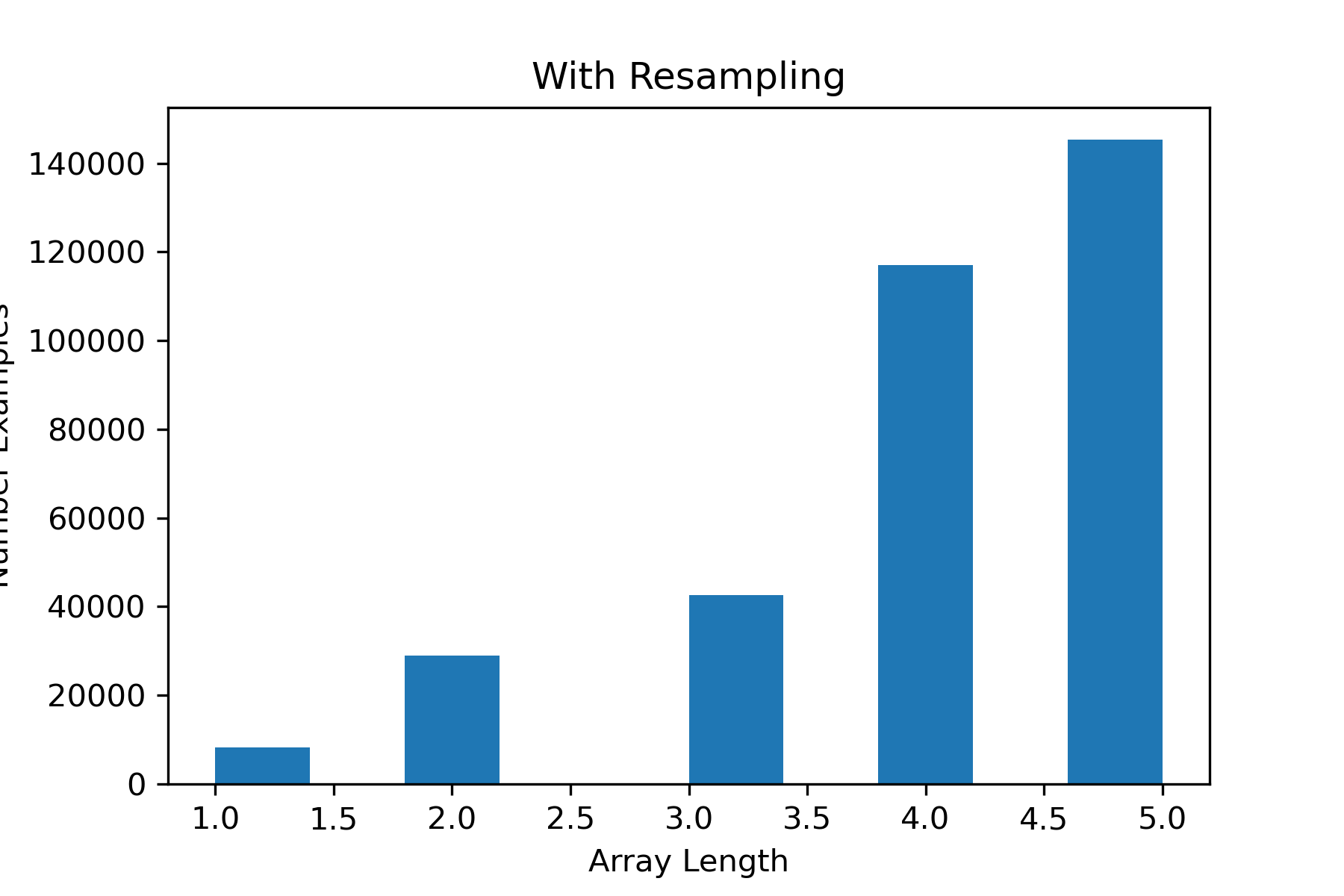}
        \caption{Dynamic programming (after resampling).}
    \end{subfigure}
    \hfill
    \begin{subfigure}[t]{0.32\textwidth}
        \centering
        \includegraphics[width=\textwidth]{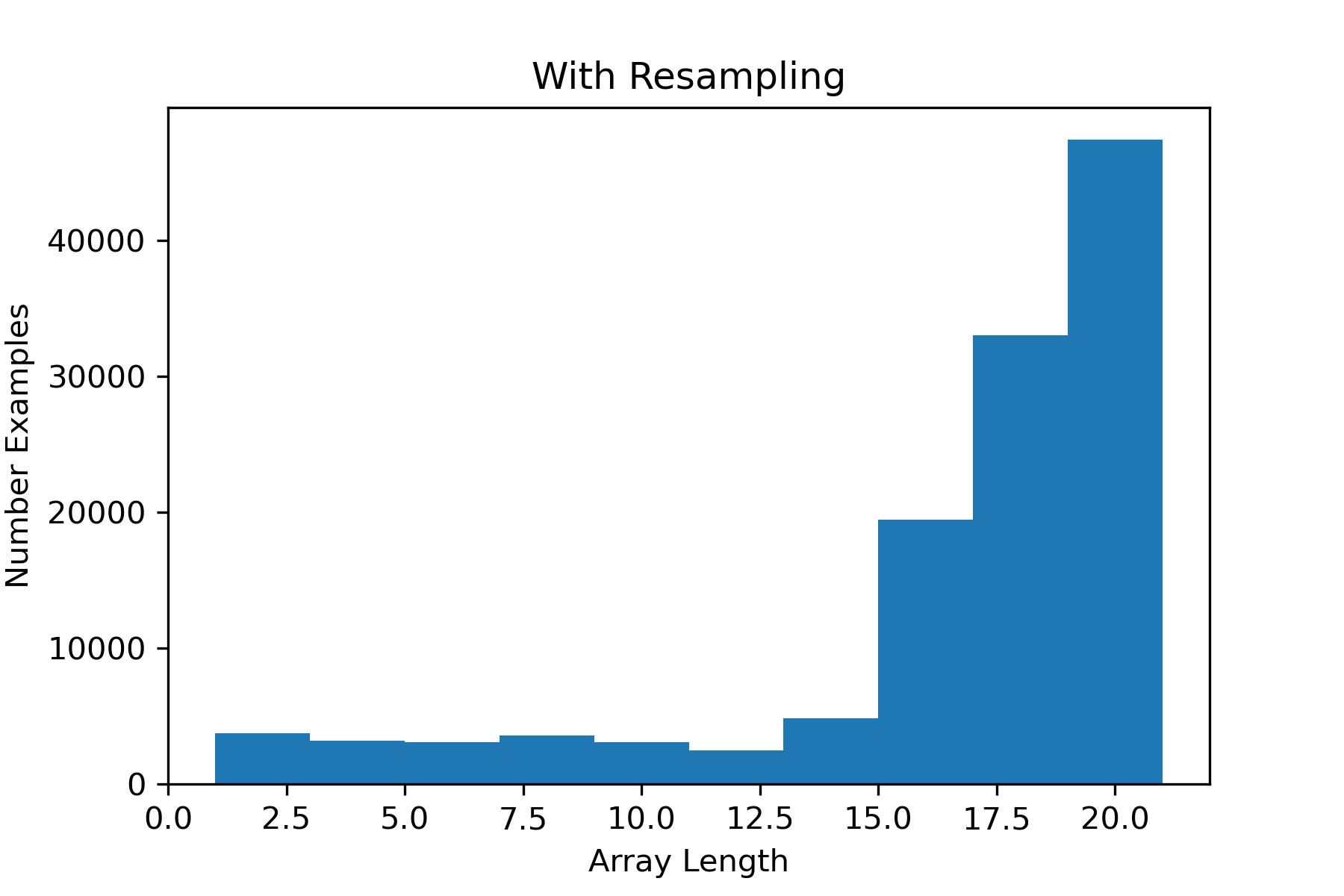}
        \caption{Parity (after resampling).}
    \end{subfigure}
    \caption{Comparison of the training dataset distributions before (top) and after (bottom) resampling on integer addition (left), dynamic programming (middle), and parity (right).}
    \label{fig:resampling_comparison}
\end{figure*}

\paragraph{Resampling} In general, we upsample examples with smaller lengths and downsample those with larger lengths in our training data. There are 2 reasons for this. First, examples with larger lengths are more numerous than examples with smaller lengths (there are many more examples of adding 2 15-digit numbers than there are adding 2 1-digit numbers). Second, since \method\ generates calls to all examples except the base case, it has trouble learning what to do in the base case if there are not enough examples. Without resampling, the base cases for each problem would be far less than 1\% of the training data. We do not follow any specific methodology for resampling. We simply try to bring the training data distribution closer to uniform than it would be without resampling. Figure~\ref{fig:resampling_comparison} displays the distributions of the training data before and after resampling with respect to length for the integer addition, dynamic programming, and parity tasks.

\paragraph{Training dataset sizes} After resampling, the final training datasets contain the following number of instances: 3,676,055 for integer addition, 342,187 for dynamic programming, and 124,780 for parity.

\paragraph{Validation and testing datasets} For each problem, we generate 5 and 100 instances of seed data for a variety of problem lengths (both in-distribution and out-of-distribution) for the validation and testing splits respectively. 

\subsection{Training Details}
\label{sec:training_details}
Training and evaluation were done on NVIDIA H100, A100, and RTX A6000 GPUs, depending on the compute requirements of the job. Rather than train the full model, we train using low-rank adapters~\citep{hu2022lora}. Hyperparameters for training jobs are in Table~\ref{table:hyperparameters}. These hyperparameters apply to training all models across all tasks, with two notable exceptions: (1) when training parity baselines, we used a slightly higher learning rate of 5e-4 for better stability, and (2) the scratchpad training job for the dynamic programming problem on LLaMA 13B used a batch size of 64, along with 64 gradient accumulation steps, so that the training job could be done on a single A100 GPU. Our training code is a heavily modified version of the code from \citet{rafailov2023direct}.
\begin{table}[h]
  \centering
  \small
  \begin{tabular}{ll}
    \toprule
    \textbf{Parameter} & \textbf{Value} \\
    \midrule
    Learning Rate    & $2 \times 10^{-4}$ \\
    LR Schedule     & Constant \\
    Optimizer        & AdamW \\
    Batch Size       & 128 \\
    Gradient accumulation steps & 32\\
    Lora\_r          & 64 \\
    Lora\_alpha      & 64 \\
    Lora\_dropout    & 0.05 \\
    \bottomrule
  \end{tabular}
  \caption{Hyperparameters used for finetuning.}
  \label{table:hyperparameters}
\end{table}

During training, cross-entropy loss is computed only on the parts of the sequence that the model will generate at inference time (c.f. grey vs. green highlighted text in the upper right of Figure~\ref{fig:main}).

We train on 500,000 samples, performing multiple epochs if necessary, and checkpoint the state model at fixed intervals. We evaluate a handful of these checkpoints on a small validation set containing 5 examples from a handful of problem lengths (both in-distribution and out-of-distribution). We report full results on the checkpoint that achieves the highest accuracy on the validation set, selecting the earlier checkpoint in the event of a tie. Table~\ref{table:samples_seen} and provide the number of examples trained on for each task (integer addition, dynamic programming, and parity) and method (baseline, scratchpad, and \method) for the selected LLaMA 7B and LLaMA 13B models.
\begin{table}
    \centering
    \small
    \begin{subtable}[t]{\linewidth} 
        \centering
            \begin{tabular}{lccc}
            \toprule
            Algorithm & \method\ & Scratchpad & Baseline \\ \midrule
            Addition & 71424 & 79360 & 349184 \\
            Parity & 63488 & 71424 & 206336 \\
            DP & 79360 & 71424 & 23808 \\
            \bottomrule
        \end{tabular}
        \label{table:samples_seen_7b}
        \caption{LLaMA 7B.}
    \end{subtable}\\
    \par\bigskip
    \begin{subtable}[t]{\linewidth}
        \centering
            \begin{tabular}{lccc}
            \toprule
            Algorithm & \method\ & Scratchpad & Baseline \\ \midrule
            Addition & 79360 & 103168 & 349184\\
            Parity & 404736 & 79360 & 206336 \\
            DP & 404736 & 104000 & 23808 \\
            \bottomrule
        \end{tabular}
        \label{table:samples_seen_13b}
        \caption{LLaMA 13B.}
    \end{subtable}
    \caption{Number of samples the selected checkpoints were trained on for LLaMA 7B (top) and LLaMA 13B (bottom).}
    \label{table:samples_seen}
\end{table}

\subsection{Inference Pipeline}
For baseline and scratchpad methods, our evaluation procedure is rather standard: we sample at a low temperature (0.01) and impose no additional context limitations beyond those of the models themselves, in which case the input is truncated from the left. With \method, we use a recursive wrapper around the same generation procedure, the pseudocode for which is shown in Algorithm~\ref{fig:psuedocode_generation}. 

To better understand the recursive generation procedure of \method, let's consider the following integer addition example: ``687 + 891\textbackslash nSolution: ''. With \method, the model is trained to return the following subproblem call ``Call: 87 + 91\textbackslash n''. In this case, we would extract the text “87 + 91” and prompt for the solution to this subproblem in a new context. Once we have the solution to this subproblem, it’s returned to the main context "687 + 891\textbackslash nSolution: Call: 87 + 91\textbackslash nReturn: 178\textbackslash nAnswer: ", and we again call the model to generate the final answer. 

\section{Additional Results}
\label{sec:galactica_results}
\begin{figure*}[t!]
    \centering
    \begin{subfigure}[t]{\textwidth}
        \centering
        \includegraphics[width=\textwidth]{figures/src/legend_7b.png}
    \end{subfigure}\\
    \vfill
    \begin{subfigure}[t]{0.32\textwidth}
        \centering
        \includegraphics[width=\textwidth]{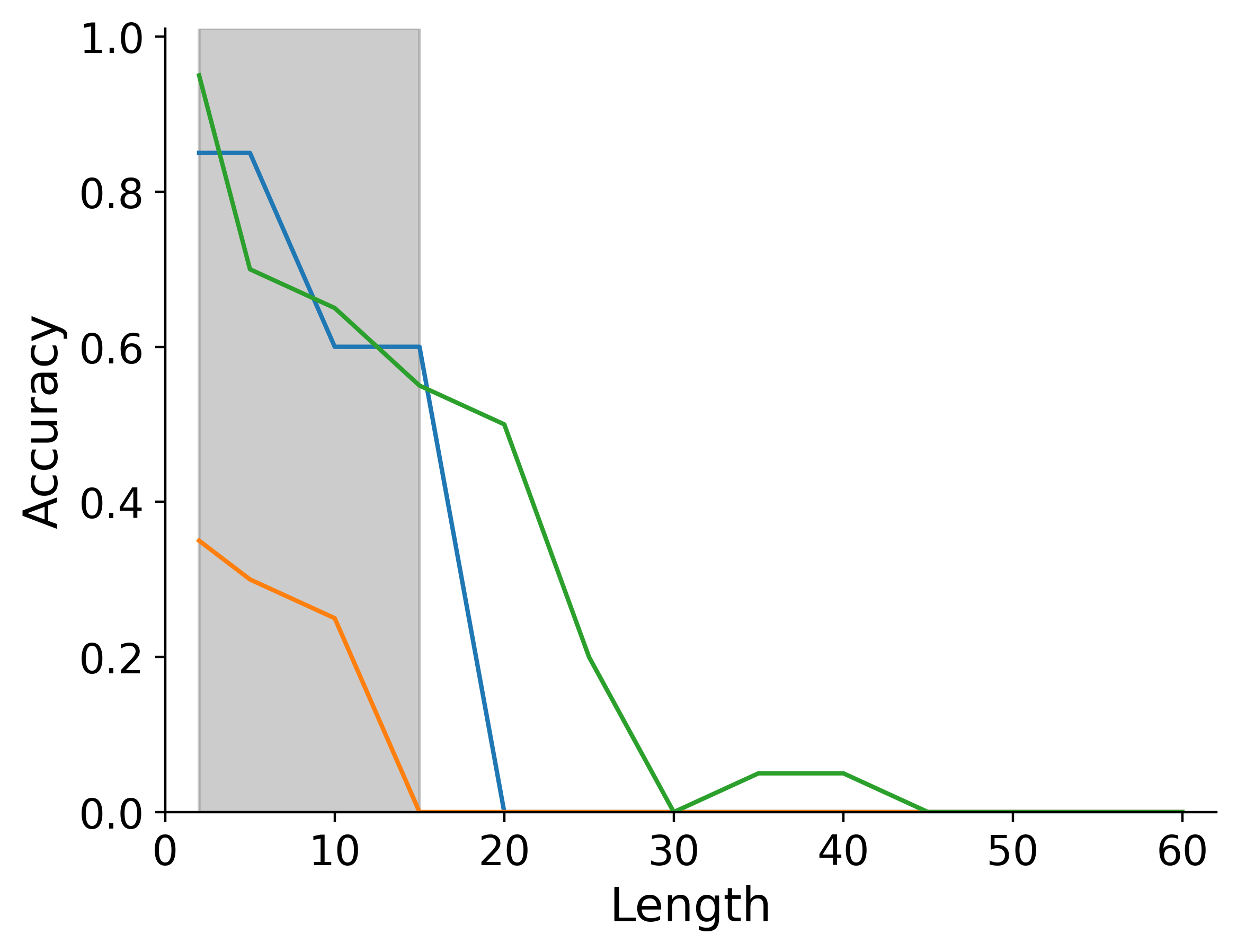}
        \caption{Galactica 125M integer addition.}
        \label{fig:addition_gal125m}
    \end{subfigure}
    \hfill
    \begin{subfigure}[t]{0.32\textwidth}
        \centering
        \includegraphics[width=\textwidth]{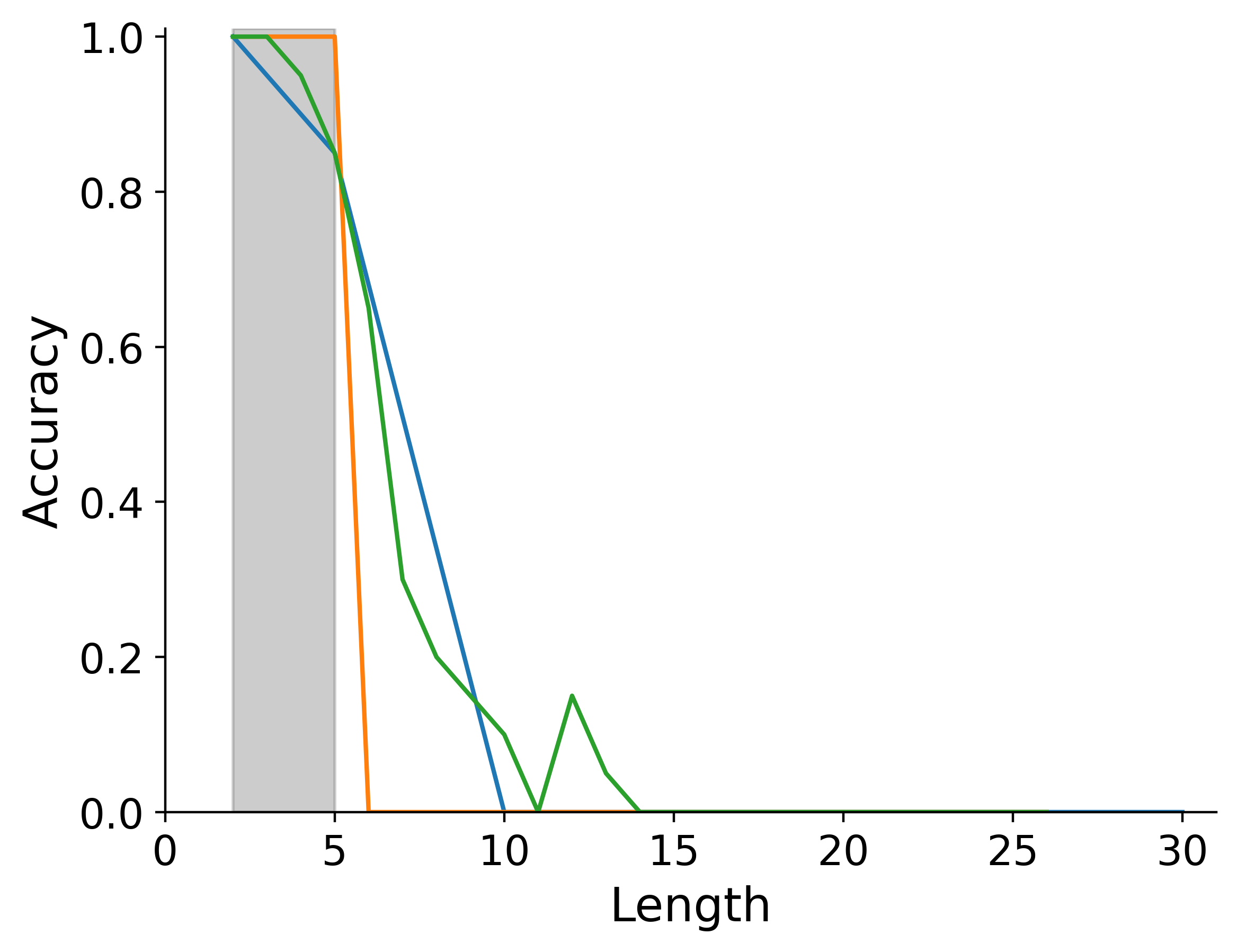}
        \caption{Galactica 125M dynamic programming.}
        \label{fig:dp_gal125m}
    \end{subfigure}
    \hfill 
    \begin{subfigure}[t]{0.32\textwidth}
        \centering
        \includegraphics[width=\textwidth]{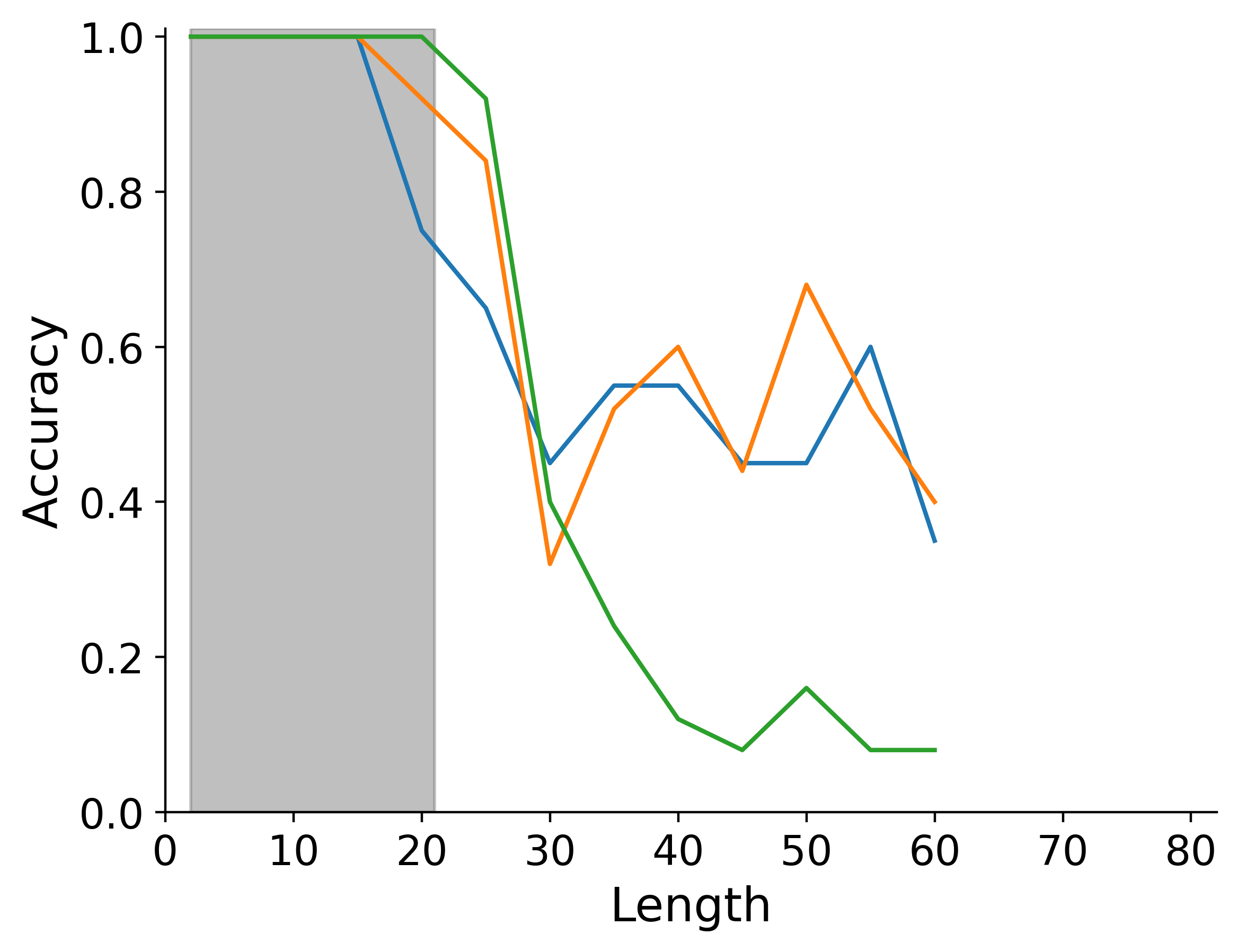}
        \caption{Galactica 125M parity.}
        \label{fig:parity_gal125m}
    \end{subfigure}\\
    \par\bigskip
    \begin{subfigure}[t]{0.32\textwidth}
        \centering
        \includegraphics[width=\textwidth]{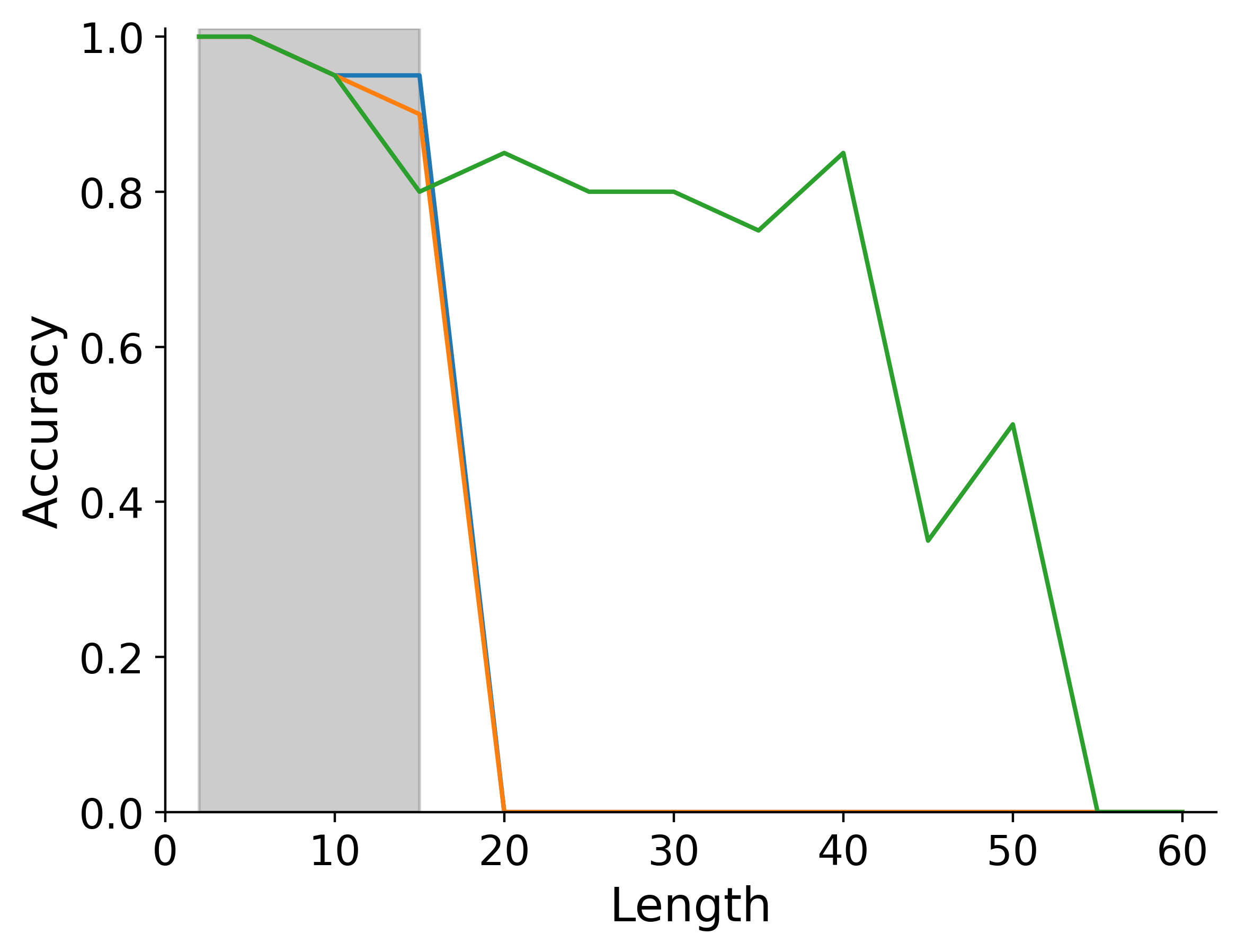}
        \caption{Galactica 1.3B integer addition.}
        \label{fig:addition_gal13b}
    \end{subfigure}
    \hfill
    \begin{subfigure}[t]{0.32\textwidth}
        \centering
        \includegraphics[width=\textwidth]{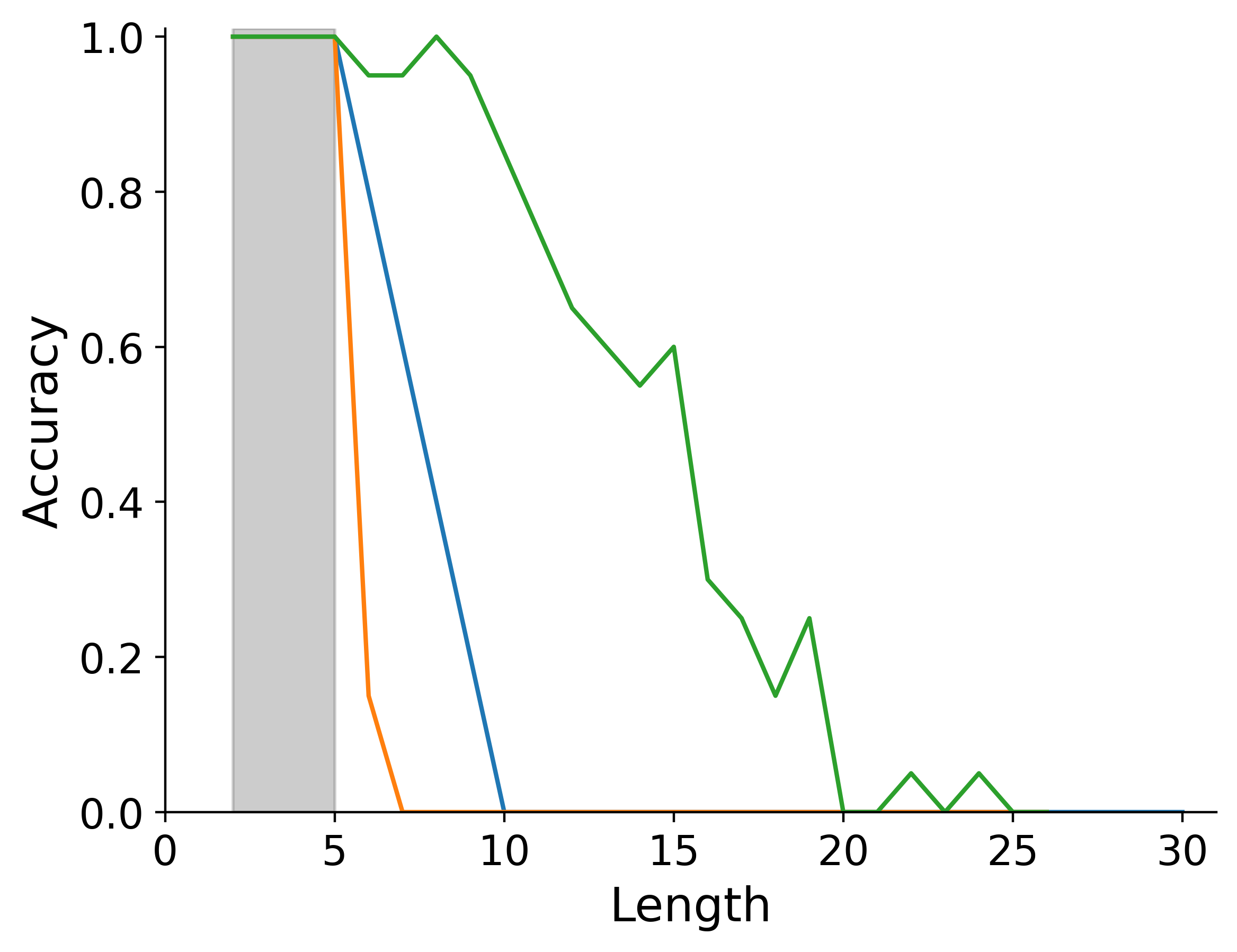}
        \caption{Galactica 1.3B dynamic programming.}
        \label{fig:dp_gal13b}
    \end{subfigure}
    \hfill 
    \begin{subfigure}[t]{0.32\textwidth}
        \centering
        \includegraphics[width=\textwidth]{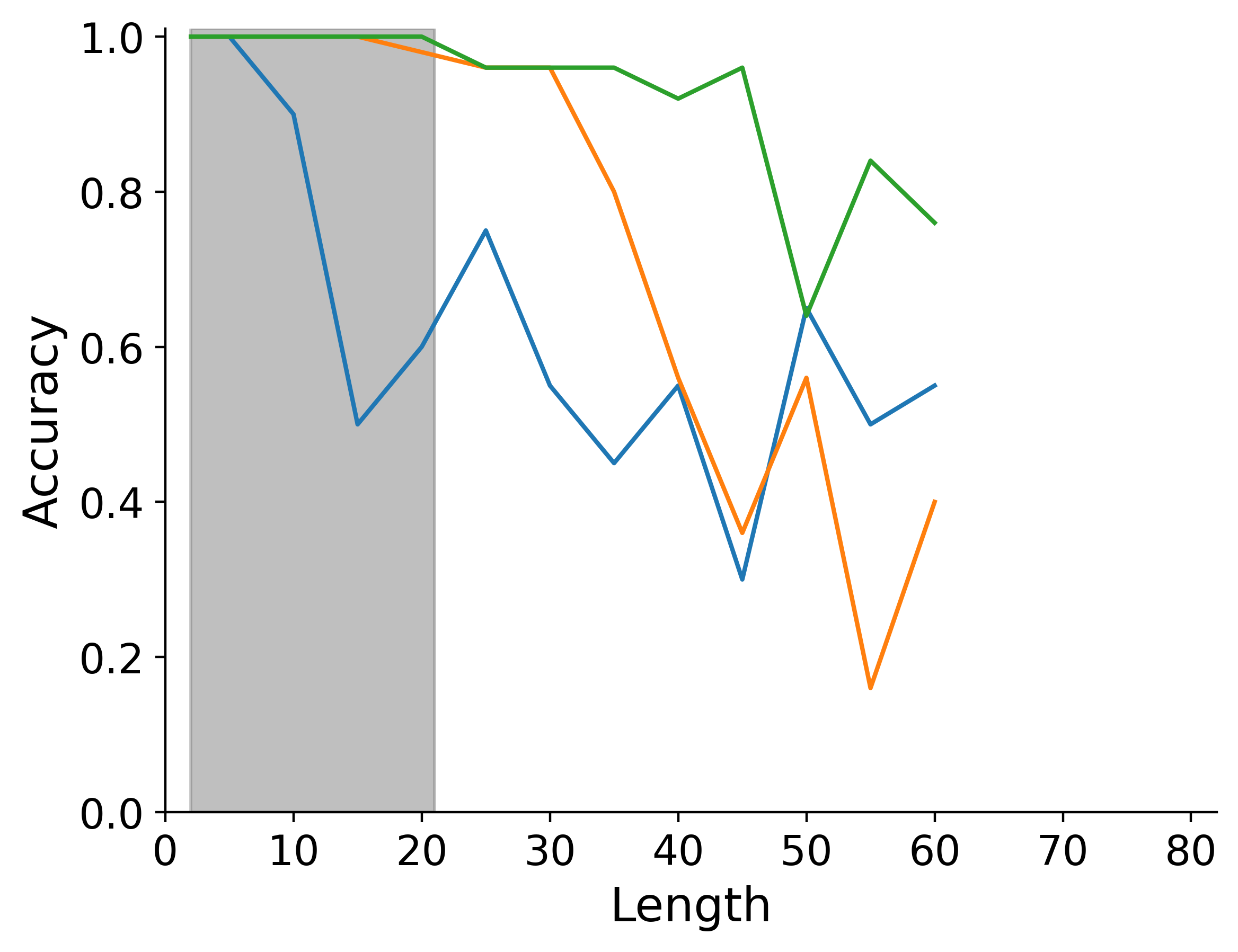}
        \caption{Galactica 1.3B parity.}
        \label{fig:parity_gal13b}
    \end{subfigure}
    \caption{Performance of Galactica 125M (top) and Galactica 1.3B (bottom) on Addition (left), Dynamic Programming (middle), and Parity (right). The in-distribution range is shaded in gray.}
    \label{fig:galactica_results}
\end{figure*}

In this section, we share results on two additional models: Galactica 125M and Galactica 1.3B~\citep{taylor2022galactica}. Results across our 3 tasks are shown in Figure~\ref{fig:galactica_results}. In general, we observe that \method\ enables Galactica 1.3B to maintain higher accuracy on more complex problem instances and Galactica 125M to perform as good, if not better, than either the baseline or scratchpad methods.

\paragraph{Integer addition} On the addition task on Galactica 125M, all three methods fail to achieve perfect in-distribution performance. With scratchpad, accuracy never tops 40\%, even for lengths seen during training, and quickly falls to 0\% at length 15. Though both baseline and \method\ methods perform similarly on in-distribution problems, \method\ generalizes better to more complex instances. With \method, Galactica 125M is able to achieve greater than 50\% accuracy at length 20, while the baseline accuracy at length 20 is 0\%. With Galactica 1.3B, again all 3 methods display similar in-distribution accuracy above 80\%. However, while the baseline and scratchpad methods are unable to solve any problems of length 20 or more correctly, Galactica 1.3B with \method\ maintains near 80\% accuracy on problems up to length 40. 

\paragraph{Dynamic Programming} On the dynamic programming task, Galactica 125M performs similarly with baseline and \method\ methods, reaching 0\% accuracy on problems of length 10 and 11 respectively. Galactica 125M with scratchpad is unable to solve any instances with out-of-distribution lengths correctly. On Galactica 1.3B, all three methods achieve perfect in-distribution performance. Still, \method\ displays better out-of-distribution performance, maintaining an accuracy near 60\% on problems of length 15, while the baseline and scratchpad methods achieve 0\% accuracy on lengths 10 and 7 respectively. Interestingly, while Galactica 1.3B with scratchpad fails to solve a single problem of length 7 correctly, Galactica 125M with \method\ can still solve above 30\% correctly, with just 1/10th of the parameters. 

\paragraph{Parity} Galactica 125M performs poorly on instances of the parity task with out-of-distribution lengths. All three methods perform similarly, and accuracy falls to that of random chance or worse on problems of length 30 or more. On Galactica 1.3B, the baseline method performs the worst, performing at the level of random chance on problems of length 15 or more. The scratchpad and \method\ maintain near-perfect performance on problems up to lengths 30 and 45 respectively.

\section{Example Problems}
\label{sec:example_problems}
In this section, we provide additional details and examples of the training data for all three tasks (integer addition, dynamic programming, and parity) with all three methods (baseline, scratchpad, and \method).

\paragraph{Integer addition} Following~\citet{liu2023goat}, we generate pairs of numbers up to 15 digits in length. With the baseline method, we train the model to output the answer directly. With scratchpad and \method, we train the model to add digits starting from the right. With these methods, the models learn to propagate the higher-order carry term separately from the rest of the output, which we found improved the accuracy of both methods. Evaluation is done on adding numbers up to 60 digits in length. To compute accuracy on testing examples, we extract the last number from the output and check for equivalence with the target solution. For the scratchpad and \method\ methods, we first prepend the carry term to the output prior to computing accuracy. Figure~\ref{fig:addition_example} displays example training instances for the baseline, scratchpad, and \method\ methods.

\paragraph{Dynamic programming} Following~\citet{dziri2023faith}, the training data for this task consists of arrays up to 5 elements long, with each element restricted to $[-5, 5]$. With the baseline method, we train the model to directly output an indices array indicating which elements should be selected to maximize the sum given the constraints. The scratchpad design is copied form~\citet{dziri2023faith}. \method\ requires two recursive calls in each context: the first to create an array with sub-array sums and the second to construct the indices array. Evaluation is done on arrays up to length 30, with each element still restricted to $[-5, 5]$. For all three methods, we extract the last array from the generated text and check for equivalence to the target indices array. Figure~\ref{fig:dp_example} displays example training instances for the baseline, scratchpad, and \method\ methods.

\paragraph{Parity} This problem was inspired by~\citet{anil2022exploring}, and requires the model to compute the parity of a binary array. Specifically, the training data includes binary arrays up to length 21, while the evaluation data includes arrays up to length 60. With the baseline method, the models learn to output the parity directly, which can be computed as the sum of the input array modulo 2. Our scratchpad design is similar to that of~\citet{anil2022exploring}, and involves keeping track of the parity sequentially as the array is traversed from left to right. With \method, the parity is computed from right to left by recursive calls made to compute the parity of the last $n-1$ elements of the array. With all three methods, we extract the last digit from the generated sequence and check for equivalence with the target parity. Figure~\ref{fig:parity_example} displays example training instances for the baseline, scratchpad, and \method\ methods.

\begin{figure*}
\centering
\small
\setlength{\fboxsep}{1em}
\setlength{\fboxrule}{1pt}
\fcolorbox{black}{white}{
\parbox{0.955\textwidth}{
\begin{alltt}
\textbf{Baseline: }\par
637 + 123\textbackslash nAnswer: \textbf{760}\par

\textbf{Scratchpad:}\par
637 + 123\textbackslash nSolution: 
\textbf{Carry 1, Output 0
\textbackslash nCarry 0, Output 60
\textbackslash nCarry 0, Output 760}\par

\textbf{\method:}\par
637 + 123\textbackslash nSolution: 
\textbf{Call: 37 + 23\textbackslash n}
Return: Carry 0, Output 60\textbackslash n
Answer: \textbf{Carry 0, Output 760}

\end{alltt}
}}
\caption{Example training instances for addition. The models learn to generate the text in bold.}
\label{fig:addition_example}
\end{figure*}

\begin{figure*}
\centering
\small
\setlength{\fboxsep}{1em}
\setlength{\fboxrule}{1pt}
\fcolorbox{black}{white}{
\parbox{0.955\textwidth}{
\begin{alltt}
\textbf{Baseline: }\par
637 + 123\textbackslash nAnswer: \textbf{760}\par
Given a sequence of integers, 
find a subsequence with the 
highest sum, such that no 
two numbers in the subsequence 
are adjacent in the original 
sequence.\textbackslash n\textbackslash nOutput a list 
with "1" for chosen numbers 
and "2" for unchosen ones. If 
multiple solutions exist, select 
the lexicographically smallest. 
Input = [1, -3, 2].\textbackslash n
\textbf{Answer: [1, 2, 1]}

\par\textbf{Scratchpad: }\par
Question: Let's solve input = [1, -3, 2]. 
\textbf{Scratchpad: dp[2] = max(input[2], 0) = max(2, 0) = 2\textbackslash ndp[1] = max(input[1], input[2], 0) = max(-3, 2, 0) = 2\textbackslash ndp[0] = max(dp[1], input[0] + dp[2], 0) = max(2, 1 + 2, 0) = 3\textbackslash n\textbackslash nFinally, we reconstruct the lexicographically smallest subsequence that fulfills the task objective by selecting numbers as follows. We store the result on a list named "output".\textbackslash n\textbackslash nLet can\_use\_next\_item = True.\textbackslash nSince dp[0] == input[0] + dp[2] (3 == 1 + 2) and can\_use\_next\_item == True, we store output[0] = 1. We update can\_use\_next\_item = False.\textbackslash nSince dp[1] != input[1] (2 != -3) or can\_use\_next\_item == False, we store output[1] = 2. We update can\_use\_next\_item = True.\textbackslash nSince dp[2] == input[2] (2 == 2) and can\_use\_next\_item == True, we store output[2] = 1.\textbackslash n\textbackslash nReconstructing all together, output=[1, 2, 1].}

\par\textbf{\method: }\par
Compute the maximum sum of nonadjacent subsequences of [1, -3, 2]\textbackslash nSolution: 
\textbf{Call: Create dp array [1, -3, 2]\textbackslash n}
Return: [3, 2, 2]\textbackslash nAnswer: 
Create chosen indices array: sum array [3, 2, 2], item array [1, -3, 2], can use item True\textbackslash nSolution: If there is only 1 item, return 1 if we should use it else 2. If we should use the first item to get the sum, call False else True. \textbf{Call: Create chosen indices array: sum array [2, 2], item array [-3, 2], can use item False\textbackslash n}Return [2, 1]\textbackslash nAnswer: Append 1 if False else 2.\textbackslash nAnswer: \textbf{[1, 2, 1]}

\end{alltt}
}}
\caption{Example training instances for dynamic programming. The models learns to generate the text in bold.}
\label{fig:dp_example}
\end{figure*}

\begin{figure*}
\centering
\small
\setlength{\fboxsep}{1em}
\setlength{\fboxrule}{1pt}
\fcolorbox{black}{white}{
\parbox{0.955\textwidth}{
\begin{alltt}
\textbf{Baseline: }\par
What is the parity of [1, 0, 1]?\textbackslash nAnswer: \textbf{0}

\par\textbf{Scratchpad}\par
What is the parity of [1, 0, 1]?
\textbackslash nSolution: Compute one element at a time
\textbf{1 1 0}

\par\textbf{\method: }\par
What is the parity of [1, 0, 1]?\textbackslash nSolution: 
\textbf{Call: What is the parity of [0, 1]?\textbackslash n}\textbackslash nReturn: 1\textbackslash nAnswer: \textbf{0}
\end{alltt}
}}
\caption{Example training instances for parity. The models learns to generate the text in bold.}
\label{fig:parity_example}
\end{figure*}
\end{document}